\documentclass[10pt,twocolumn,letterpaper]{article}

\usepackage{iccv}
\usepackage{times}

\usepackage{graphicx}
\usepackage{booktabs} % for professional tables
\usepackage[pagebackref=true,breaklinks=true,colorlinks,bookmarks=false]{hyperref}
\usepackage{xcolor}
\usepackage{url}            % simple URL typesetting
\usepackage{amsfonts}       % blackboard math symbols
\usepackage{nicefrac}       % compact symbols for 1/2, etc.
\usepackage{microtype}      % microtypography
\usepackage{subcaption}
\usepackage{multirow}
\usepackage{amsmath}
\usepackage{amsthm}
\usepackage{mathtools}
\usepackage{verbatim}
\usepackage{amssymb}
\usepackage{colortbl}
\usepackage{xspace}
\usepackage{tikz}
\usepackage{afterpage}
\usepackage{enumitem}
\usepackage[square,numbers,sort&compress]{natbib}
\renewcommand{\citealp}[1]{\citet{#1}}
\let\oldciteauthor=\citeauthor
\renewcommand{\citeauthor}[1]{{\protect\NoHyper\oldciteauthor{#1}\protect\endNoHyper}\xspace}

\usepackage{caption}
\newcommand{\One}{\mathbf{1}}

\DeclareMathOperator*{\argmax}{argmax}
\newcommand{\mnist}{\textsc{Mnist}\xspace}
\newcommand{\cifar}{\textsc{Cifar-10}\xspace}
\newcommand{\svhn}{\textsc{Svhn}\xspace}
\newcommand{\imagenet}{\textsc{ImageNet}\xspace}
\newcommand{\wideresnet}{WideResNet\xspace}

\newcommand{\linf}{\ensuremath{\ell_\infty}\xspace}
\newcommand{\T}{^\textsf{T}}
\newcommand{\y}{\ensuremath{y_\textrm{true}}\xspace}
\newcommand{\onehot}[1]{\ensuremath{e_{#1}}\xspace}
\DeclareMathOperator*{\maximize}{max}
\DeclareMathOperator*{\minimize}{min}
\newcommand{\lbound}[2]{\ensuremath{\underline{#1}_{#2}}\xspace}
\newcommand{\ubound}[2]{\ensuremath{\overline{#1}_{#2}}\xspace}
\newcommand{\lowerbound}[2]{\ensuremath{\lbound{#1}{#2}(\epsilon)}\xspace}
\newcommand{\upperbound}[2]{\ensuremath{\ubound{#1}{#2}(\epsilon)}\xspace}
\newcommand{\epsilontrain}{\ensuremath{\epsilon_{\textrm{train}}}\xspace}

\definecolor{Highlight}{gray}{0.9}

\makeatletter
\def\bstctlcite{\@ifnextchar[{\@bstctlcite}{\@bstctlcite[@auxout]}}
\def\@bstctlcite[#1]#2{\@bsphack
  \@for\@citeb:=#2\do{%
    \edef\@citeb{\expandafter\@firstofone\@citeb}%
    \if@filesw\immediate\write\csname #1\endcsname{\string\citation{\@citeb}}\fi}%
  \@esphack}
\makeatother

\iccvfinalcopy % *** Uncomment this line for the final submission
\newcommand{\appendixref}[2]{\ifx\cutpaper\undefined\ref{#1}\else#2\fi}

\newcommand{\arxiv}{}  % *** Uncomment to get the ArXiv version.
\ifx\arxiv\undefined
\else
\let\cutpaper\undefined
\fi

\def\iccvPaperID{4660} % *** Enter the ICCV Paper ID here

\ificcvfinal\fi

\begin{document}
\bstctlcite{BSTcontrol}  % Bibliography magic.

\title{
\ifx\arxiv\undefined
Scalable Verified Training for Provably Robust Image Classification
\else
On the Effectiveness of Interval Bound Propagation for\\Training Verifiably Robust Models
\fi
}

\author{Sven Gowal\textsuperscript{*}\\
DeepMind\\
{\tt\small sgowal@google.com}
\and
Krishnamurthy (Dj) Dvijotham\textsuperscript{*}\\
% DeepMind\\
{\tt\small dvij@google.com}
\and
Robert Stanforth\textsuperscript{*}\\
% DeepMind\\
{\tt\small stanforth@google.com}
\and
Rudy Bunel\\
% University of Oxford\\
% {\tt\small rudy@robots.ox.ac.uk}
\and
Chongli Qin\\
% DeepMind\\
% {\tt\small chongliqin@google.com}
\and
Jonathan Uesato\\
% DeepMind\\
% {\tt\small juesata@google.com}
\and
Relja Arandjelovi\'c\\
% DeepMind\\
% {\tt\small relja@google.com}
\and
Timothy Mann\\
% DeepMind\\
% {\tt\small timothymann@google.com}
\and
Pushmeet Kohli\\
% DeepMind\\
% {\tt\small pushmeet@google.com}
}
\maketitle

\begin{abstract}
Recent work has shown that it is possible to train deep neural networks that are provably robust to norm-bounded adversarial perturbations.
Most of these methods are based on minimizing an upper bound on the worst-case loss over all possible adversarial perturbations.
While these techniques show promise, they often result in difficult optimization procedures that remain hard to scale to larger networks.
Through a comprehensive analysis, we show how a simple bounding technique, interval bound propagation (IBP), can be exploited to train large provably robust neural networks that beat the state-of-the-art in verified accuracy.
While the upper bound computed by IBP can be quite weak for general networks, we demonstrate that an appropriate loss and clever hyper-parameter schedule allow the network to adapt such that the IBP bound is tight.
This results in a fast and stable learning algorithm that outperforms more sophisticated methods and achieves state-of-the-art results on \mnist, \cifar and \svhn.
It also allows us to train the largest model to be verified beyond vacuous bounds on a downscaled version of \imagenet.
\end{abstract}

\section{Introduction}

\begin{figure}[t]
\centering
\begin{footnotesize}
\begin{tikzpicture}[scale=0.95]
    \node[anchor=south west,inner sep=0,rotate=0] (image) at (0,0) {\includegraphics[width=.98\linewidth]{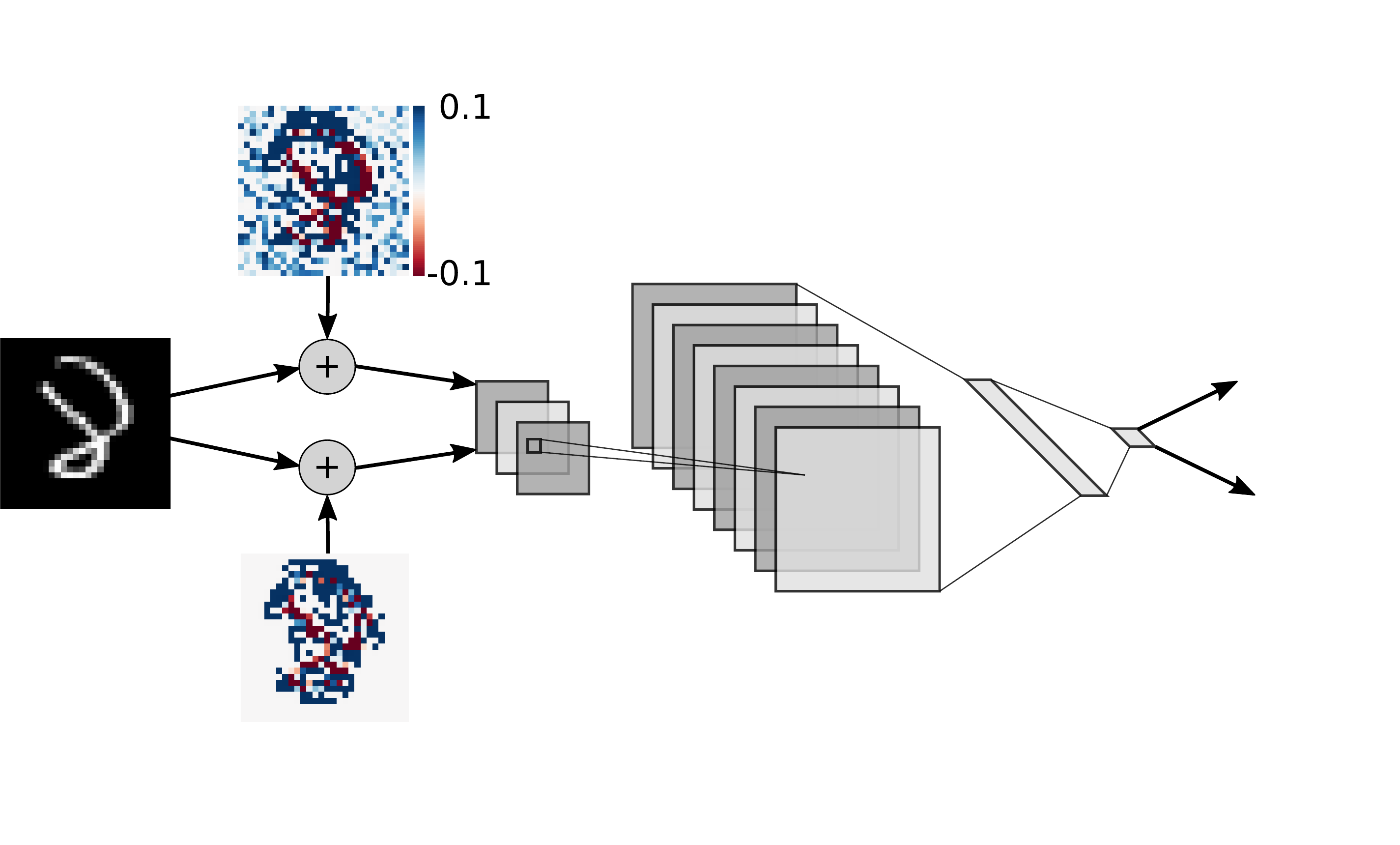}};
    \begin{scope}[x={(image.south east)},y={(image.north west)}]
        \draw (.23, .15) node [text width=3cm,align=center,anchor=north] {Perturbation found\\through exhaustive search\\(via MIP solver)};
        \draw (.23, 1.05) node [text width=3cm,align=center,anchor=north] {Perturbation found\\using PGD};
        
        \draw (.58, .85) node [text width=4cm,align=center,anchor=north] {Empirically robust but\\not provably robust model};
        
        \draw (0.9, 0.85) node [text width=3cm,align=center,anchor=north] {Predictions};
        \draw (.89, .41) node [text width=3cm,align=left,anchor=west] {2 (MIP)};
        \draw (.88, .56) node [text width=3cm,align=left,anchor=west] {8 (PGD)};
        
        % Help lines to place annotations (to comment after usage)
        %\draw[help lines,xstep=.1,ystep=.1] (0,0) grid (1,1);
        %\foreach \x in {0,1,...,9} { \node [anchor=north] at (\x/10,0) {0.\x}; }
        %\foreach \y in {0,1,...,9} { \node [anchor=east] at (0,\y/10) {0.\y}; }
    \end{scope}
    % \draw (current bounding box.north east) -- (current bounding box.north west) -- (current bounding box.south west) -- (current bounding box.south east) -- cycle;
\end{tikzpicture}
\end{footnotesize}
\caption{Example motivating why robustness to the projected gradient descent (PGD) attack is not a true measure of robustness (even for small convolutional neural networks). Given a seemingly robust neural network, the worst-case perturbation of size $\epsilon = 0.1$ found using 200 PGD iterations and 10 random restarts (shown at the top) is correctly classified as an ``eight''. However, a worst case perturbation classified as a ``two'' can be found through exhaustive search (shown at the bottom).}
\label{fig:problem}
\end{figure}
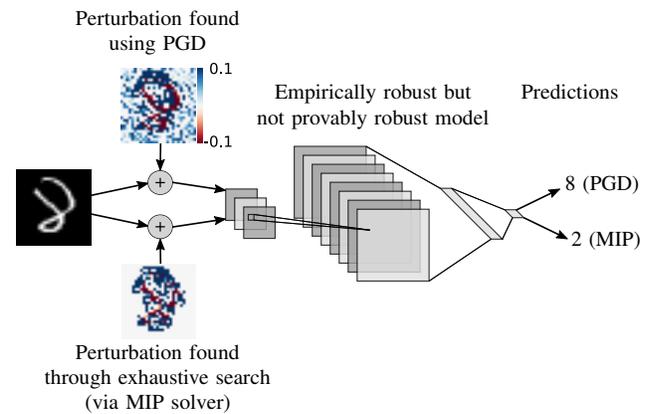

Despite the successes of deep learning \citep{goodfellow2016deep}, it is well-known that neural networks are not robust.
In particular, it has been shown that the addition of small but carefully chosen deviations to the input, called adversarial perturbations, can cause the neural network to make incorrect predictions with high confidence \citep{carlini2017adversarial,carlini2017towards,goodfellow2014explaining,kurakin2016adversarial,szegedy2013intriguing}.
Starting with \citet{szegedy2013intriguing}, there has been a lot of work on understanding and generating adversarial perturbations \citep{carlini2017towards,athalye2017synthesizing}, and on building models that are robust to such perturbations \citep{goodfellow2014explaining,papernot2015distillation,madry2017towards,kannan2018adversarial}.
Unfortunately, many of the defense strategies proposed in the literature are targeted towards a specific adversary (e.g., obfuscating gradients against projected gradient attacks), and as such they are easily broken by stronger adversaries \citep{uesato2018adversarial,athalye2018obfuscated}.
Robust optimization techniques, like the one developed by \citet{madry2017towards}, overcome this problem by trying to find the worst-case adversarial examples at each training step and adding them to the training data.
While the resulting models show strong empirical evidence that they are robust against many attacks, we cannot yet guarantee that a different adversary (for example, one that does brute-force enumeration to compute adversarial perturbations) cannot find inputs that cause the model to predict incorrectly.
In fact, Figure~\ref{fig:problem} provides an example that motivates why projected gradient descent (PGD) -- the technique at the core of \citeauthor{madry2017towards}'s method -- does not always find the worst-case attack (a phenomenon also observed by \citealp{tjeng2017verifying}).

This has driven the need for \emph{formal verification}: a provable guarantee that neural networks are consistent with a \emph{specification} for all possible inputs to the network.
Substantial progress has been made: from complete methods that use Satisfiability Modulo Theory (SMT) \citep{katz2017reluplex,ehlers2017formal,carlini2017ground} or Mixed-Integer Programming (MIP)  \citep{bunel2017piecewise,tjeng2017verifying,cheng2017maximum}  to incomplete methods that rely on solving a convex \emph{relaxation} of the verification problem \citep{weng2018towards,mirman2018differentiable,gehr2018ai,dvijotham2018dual,dvijotham2018training,wong2018provable,wong2018scaling,wang2018formal}.
Complete methods, which provide exact robustness bounds, are expensive and difficult to scale (since they perform exhaustive enumeration in the worst case).
Incomplete methods provide robustness bounds that can be loose.
However, they scale to larger models than complete methods and, as such, can be used inside the training loop to build models that are not only robust, but also intrinsically easier to verify \citep{raghunathan2018certified,wong2018provable,mirman2018differentiable,dvijotham2018training}.

In this paper, we study interval bound propagation (IBP), which is derived from interval arithmetic \citep{sunaga1958theory,katz2017reluplex,ehlers2017formal}: an incomplete method for training verifiably robust classifiers.
IBP allows to define a loss  to minimize an upper bound on the maximum difference between any pair of logits when the input can be perturbed within an \linf norm-bounded ball.
Compared to more sophisticated approaches \citep{raghunathan2018certified, wong2018provable, mirman2018differentiable, dvijotham2018training}, IBP is very fast -- its computational cost is comparable to two forward passes through the network.
This enables us to have a much faster training step, allowing us to scale to larger models with larger batch sizes and perform more extensive hyper-parameter search.
While the core approach behind IBP has been studied to some extent in previous papers \citep{dvijotham2018training,mirman2018differentiable}, blindly using it results in a difficult optimization problem with unstable performance.
Most notably, we develop a training curriculum and show that this approach can achieve strong results, outperforming the state-of-the-art.
%The core approach behind IBP has been studied in previous papers -- it is equivalent to the \emph{constant verifier} used by \citet{dvijotham2018training} and to the \emph{interval domain} from \citet{mirman2018differentiable}.
The contributions of this paper are as follows:
\begin{itemize}[topsep=0pt,itemsep=2pt,partopsep=1pt, parsep=1pt]
    \item We propose several enhancements that improve the performance of IBP for verified training.
    In particular, we differentiate ourselves from~\citet{mirman2018differentiable} by using a different loss function, and by eliding the last linear layer of the neural network, thereby improving our estimate of the worst-case logits.
    We also develop a curriculum that stabilizes training and improves generalization.
    % We explain our training methodology by detailing how key hyper-parameters are scheduled throughout training.
    \item We compare our trained models to those from other approaches in terms of robustness to PGD attacks~\citep{carlini2017towards} and show that they are competitive against \citet{madry2017towards} and \citet{wong2018scaling} across a wide range of \linf perturbation radii (hereafter denoted by $\epsilon$).
    We also compare IBP to \citeauthor{wong2018scaling}'s method in terms of verified error rates.
    % by using a complete Mixed-Integer Programming (MIP) solver.
    \item We demonstrate that IBP is not only computationally cheaper, but that it also achieves the state-of-the-art verified accuracy for single-model architecture.\footnote{The use of ensembles or cascades (as done by \citealp{wong2018scaling}) is orthogonal to the work presented here.}
    We reduce the verified error rate from 3.67\% to 2.23\% on \mnist (with \linf perturbations of $\epsilon = 0.1$\footnote{$\epsilon$ is measured with respect to images normalized between 0 and 1.}), from 19.32\% to 8.05\% on \mnist (at $\epsilon = 0.3$), and from 78.22\% to 67.96\% on \cifar (at $\epsilon = 8/255$).
    Thus, demonstrating the extent to which the model is able to adapt itself during training so that the simple relaxation induced by IBP is not too weak.
    \item We train the first provably robust model on \imagenet (downscaled to $64 \times 64$ images) at $\epsilon = 1/255$. Using a \wideresnet-10-10, we reach 93.87\% top-1 verified error rate. This constitutes the largest model to be verified beyond vacuous bounds (a random or constant classifier would achieve a 99.9\% verified error rate).
    \item Finally, the code for training provably robust neural networks using IBP is available at {\footnotesize\url{https://github.com/deepmind/interval-bound-propagation}}.
\end{itemize}

\section{Related Work}\label{sec:related}

Work on training verifiably robust neural networks typically falls in one of two primary categories.
First, there are empirical approaches exemplified perfectly by \citet{xiao2018training}.
This work takes advantage of the nature of MIP-based verification -- the critical bottleneck being the number of integer variables the solver needs to branch over.
The authors design a regularizer that aims to reduce the number of ambiguous ReLU activation units (units for which bound propagation is not able to determine whether they are on or off) so that verification after training using a MIP solver is efficient.
This method, while not providing any meaningful measure of the underlying verified accuracy during training, is able to reach state-of-the-art performance once verified after training with a MIP solver.

Second, there are methods that compute a differentiable upper bound on the violation of the specification to verify.
This upper bound, if fast to compute, can be used within a loss (e.g., hinge loss) to optimize models through regular Stochastic Gradient Descent (SGD).
In this category, we highlight the works by \citet{raghunathan2018certified}, \citet{wong2018scaling}, \citet{dvijotham2018training} and \citet{mirman2018differentiable}.
\citet{raghunathan2018certified} use a semi-definite relaxation that provides an adaptive regularizer that encourages robustness.
\citet{wong2018scaling} extend their previous work \cite{wong2018provable}, which considers the dual formulation of the underlying LP.
Critically, any feasible dual solution provides a guaranteed upper bound on the solution of the primal problem.
This allows \citeauthor{wong2018provable} to fix the dual solution and focus on computing tight activation bounds that, in turn, yield a tight upper bound on the specification violation.
Alternatively, \citet{dvijotham2018training} fix the activation bounds and optimize the dual solution using an additional \emph{verifier} network.
Finally, \citet{mirman2018differentiable} introduce geometric abstractions that bound activations as they propagate through the network.
% These convex abstractions, called \emph{domains}, are not only differentiable, but they are also computationally efficient to propagate.
To the contrary of the conclusions from these previous works, we demonstrate that tighter relaxations (such as the dual formulation from \citet{dvijotham2018training}, or the \emph{zonotope domain} from \citet{mirman2018differentiable}) are not necessary to reach tight verified bounds.
% As previously mentioned, IBP is equivalent to the \emph{constant verifier} used by \citet{dvijotham2018training} and to the \emph{interval domain} from \citet{mirman2018differentiable}.

IBP, which often leads to loose upper bounds for arbitrary networks, has a significant computational advantage, since computing IBP bounds only requires two forward passes through the network.
This enables us to apply IBP to significantly larger models and train with extensive hyper-parameter tuning.
We show that thanks to this capability, a carefully tuned verified training process using IBP is able to achieve state-of-the-art verified accuracy.
Perhaps surprisingly, our results show that neural networks can easily adapt to make the rather loose bound provided by IBP much tighter -- this is in contrast to previous results that seemed to indicate that more expensive verification procedures are needed to improve the verified accuracy of neural networks in image classification tasks.

\section{Methodology}\label{sec:ibp}

\paragraph{Neural network.}

We focus on feed-forward neural networks trained for classification tasks.
The input to the network is denoted $x_0$ and its output is a vector of raw un-normalized predictions (hereafter logits) corresponding to its beliefs about which class $x_0$ belongs to.
During training, the network is fed pairs of input $x_0$ and correct output label \y, and trained to minimize a misclassification loss, such as cross-entropy.

For clarity of presentation, we assume that the neural network is defined by a sequence of transformations $h_k$ for each of its $K$ layers. That is, for an input $z_0$ (which we define formally in the next paragraph), we have
\begin{equation}
z_k = h_k(z_{k - 1}) \quad k = 1, \ldots, K
\end{equation}
The output $z_K \in \mathbb{R}^N$ has $N$ logits corresponding to $N$ classes.

\paragraph{Verification problem.}

We are interested in verifying that neural networks satisfy a specification by generating a proof that this specification holds.
We consider specifications that require that for all inputs in some set $\mathcal{X}(x_0)$ around $x_0$, the network output satisfies a linear relationship
\begin{equation}
    c\T z_K + d \leq 0 \quad \forall z_0 \in \mathcal{X}(x_0)
    \label{eq:spec}
\end{equation}
where $c$ and $d$ are a vector and a scalar that may depend on the nominal input $x_0$ and label \y.
As shown by \citet{dvijotham2018dual}, many useful verification problems fit this definition.
In this paper, we focus on the robustness to adversarial perturbations within some \linf norm-bounded ball around the nominal input $x_0$.

A network is adversarially robust at a point $x_0$ if there is no choice of adversarial perturbation that changes the classification outcome away from the true label \y, i.e., $\argmax_i z_{K,i} = \y$ for all elements $z_0 \in \mathcal{X}(x_0)$.
Formally, we want to verify that for each class $y$:
\begin{equation}
    (\onehot{y} - \onehot{\y})\T z_K \leq 0 \quad \forall z_0 \in \mathcal{X}(x_0) = \{x ~|~ \|x - x_0\|_\infty < \epsilon\}
\label{eq:adversarial_spec}
\end{equation}
where \onehot{i} is the standard $i^\textrm{th}$ basis vector and $\epsilon$ is the perturbation radius.

Verifying a specification like \eqref{eq:spec} can be done by searching for a counter-example that violates the specification constraint:
\begin{equation}
\begin{aligned}
& \underset{z_0 \in \mathcal{X}(x_0)}{\maximize} & & c\T z_K + d  \\
& \text{subject to} & & z_k = h_k(z_{k - 1}) \quad k = 1, \ldots, K
\end{aligned}
\label{eq:verification}
\end{equation}
If the optimal value of the above optimization problem is smaller than 0, the specification \eqref{eq:spec} is satisfied.

\paragraph{Interval bound propagation.}
\label{sec:nbp}

\begin{figure}[t]
\centering
\begin{footnotesize}
\begin{tikzpicture}
    \clip (0, -0.28) rectangle (8.2, 3.7);
    \node[anchor=south west,inner sep=0,rotate=0] (image) at (0,0) {\includegraphics[width=.98\linewidth]{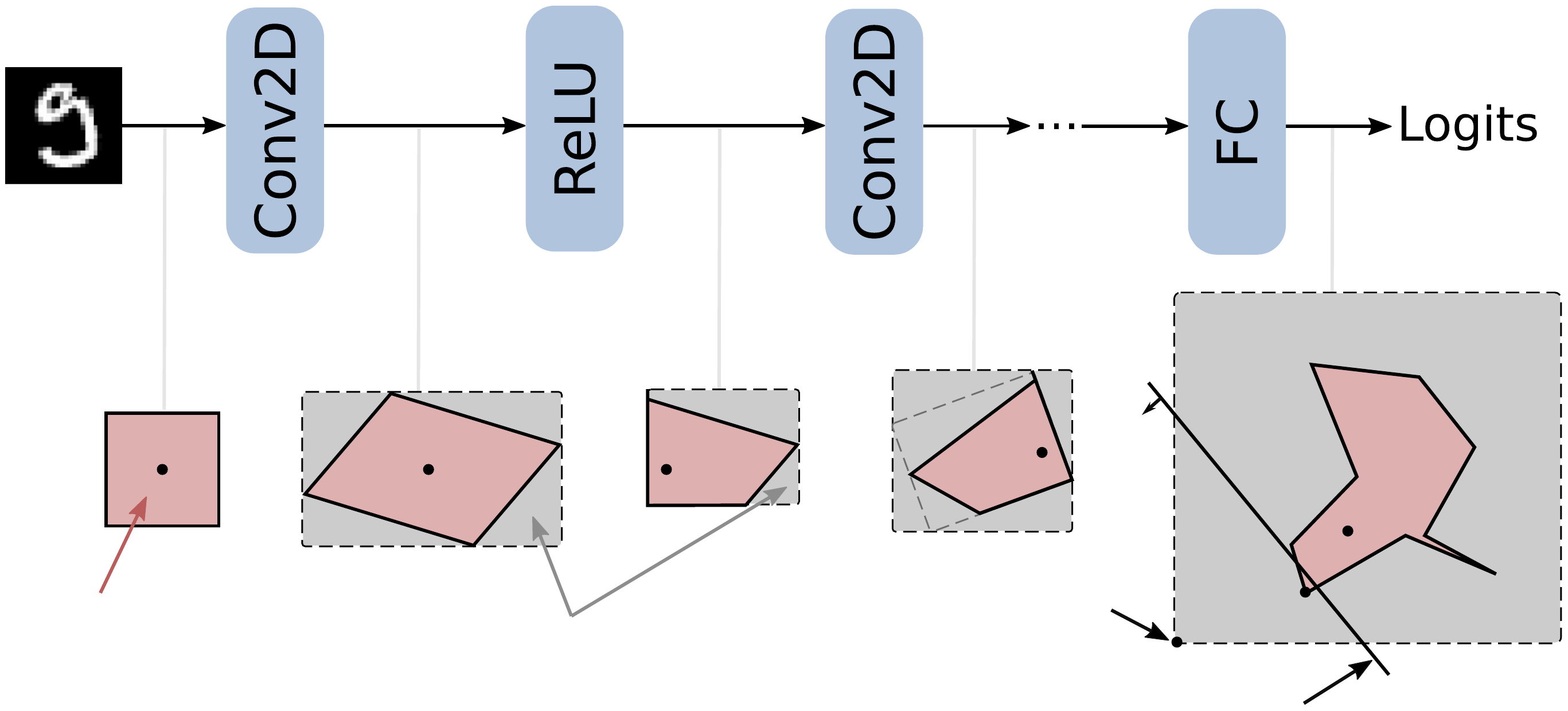}};
    \begin{scope}[x={(image.south east)},y={(image.north west)}]
        \draw (.1, .18) node [text width=3cm,align=center,anchor=north] {Adversarial\\polytope};
        \draw (.35, .15) node [text width=3cm,align=center,anchor=north] {Interval\\bounds};
        
        \draw (0.73, 0.03) node [text width=3cm,align=center,anchor=north] {Specification};
        \draw (.71, .12) node [text width=3cm,align=right,anchor=east] {Upper\\bound};
        
        % Help lines to place annotations (to comment after usage)
        %\draw[help lines,xstep=.1,ystep=.1] (0,0) grid (1,1);
        %\foreach \x in {0,1,...,9} { \node [anchor=north] at (\x/10,0) {0.\x}; }
        %\foreach \y in {0,1,...,9} { \node [anchor=east] at (0,\y/10) {0.\y}; }
    \end{scope}
    % \draw (current bounding box.north east) -- (current bounding box.north west) -- (current bounding box.south west) -- (current bounding box.south east) -- cycle;
\end{tikzpicture}
\end{footnotesize}
\caption{Illustration of interval bound propagation. From the left, the adversarial polytope (illustrated in 2D for clarity) of the nominal image of a ``nine'' (in red) is propagated through a convolutional network. At each layer, the polytope deforms itself until the last layer where it takes a complicated and non-convex shape in logit space. Interval bounds (in gray) can be propagated similarly: after each layer the bounds are reshaped to be axis-aligned bounding boxes that always encompass the adversarial polytope. In logit space, it becomes easy to compute an upper bound on the worst-case violation of the specification to verify.}
\label{fig:diagram}
\end{figure}
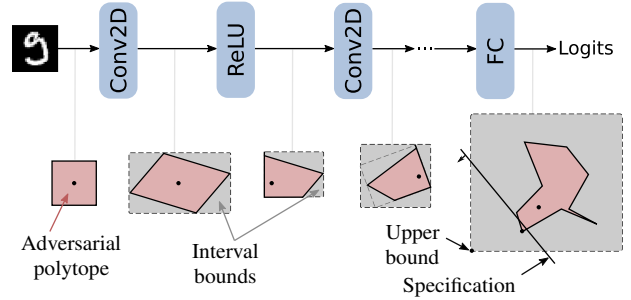

IBP's goal is to find an upper bound on the optimal value of the  problem \eqref{eq:verification}.
The simplest approach is to bound the activation $z_k$ of each layer by an axis-aligned bounding box (i.e., $\lowerbound{z}{k} \leq z_k \leq \upperbound{z}{k}$~~\footnote{For simplicity, we abuse the notation $\leq$ to mean that all coordinates from the left-hand side need to be smaller than the corresponding coordinates from the right-hand side.}) using interval arithmetic. 
For \linf adversarial perturbations of size $\epsilon$, we have for each coordinate $z_{k,i}$ of $z_k$:
\begin{equation}
\begin{aligned}
\lowerbound{z}{k,i} = \!\!\!\!\!\!\!\!\underset{\lowerbound{z}{k-1} \leq z_{k-1} \leq \upperbound{z}{k-1}}{\minimize}\!\! \onehot{i}\T h_k(z_{k - 1}) \\
\upperbound{z}{k,i} = \!\!\!\!\!\!\!\!\underset{\lowerbound{z}{k-1} \leq z_{k-1} \leq \upperbound{z}{k-1}}{\maximize}\!\! \onehot{i}\T h_k(z_{k - 1})
\end{aligned}
\label{eq:bound_propagation}
\end{equation}
where $\lowerbound{z}{0} = x_0 - \epsilon \One$ and $\upperbound{z}{0} = x_0 + \epsilon \One$.
The above optimization problems can be solved quickly and in closed form for affine layers and monotonic activation functions.
An illustration of IBP is shown in Figure~\ref{fig:diagram}.

For the \textbf{affine layers} (e.g., fully connected layers, convolutions) that can be represented in the form  $h_k(z_{k-1}) = Wz_{k-1} + b$, solving the optimization problems~\eqref{eq:bound_propagation} can be done efficiently with only two matrix multiplies:
\begin{equation}
\begin{aligned}
\mu_{k-1} &= \frac{\ubound{z}{k-1} + \lbound{z}{k-1}}{2} \\
r_{k-1} &= \frac{\ubound{z}{k-1} - \lbound{z}{k-1}}{2} \\
\mu_k &= W \mu_{k-1} + b \\
r_k &= |W| r_{k-1} \\
\lbound{z}{k} &= \mu_k - r_k \\
\ubound{z}{k} &= \mu_k + r_k
\end{aligned}
\label{eq:affine_bounds}
\end{equation}
where $|\cdot|$ is the element-wise absolute value operator. 
Propagating bounds through any element-wise \textbf{monotonic activation function} (e.g., $\textrm{ReLU}$, $\textrm{tanh}$, $\textrm{sigmoid}$) is trivial.
Concretely, if $h_k$ is an element-wise increasing function, we have:
\begin{equation}
\begin{aligned}
\lbound{z}{k} &= h_k(\lbound{z}{k-1}) \\
\ubound{z}{k} &= h_k(\ubound{z}{k-1})
\end{aligned}
\label{eq:act_bounds}
\end{equation}

Notice how for element-wise non-linearities the $(\lbound{z}{k}, \ubound{z}{k})$ formulation is better, while for affine transformations $(\mu_k, r_k)$ is more efficient (requiring two matrix multiplies instead of four).
Switching between parametrizations depending on $h_k$ incurs a slight computational overhead, but since affine layers are typically more computationally intensive, the formulation \eqref{eq:affine_bounds} is worth it.

Finally, the upper and lower bounds of the output logits $z_K$ can be used to construct an upper bound on the solution of \eqref{eq:verification}:
\begin{equation}
\underset{\lowerbound{z}{K} \leq z_K \leq \upperbound{z}{K}}{\maximize} c\T z_K + d
\label{eq:upper_bound_verification}
\end{equation}
Overall, the adversarial specification \eqref{eq:adversarial_spec} is upper-bounded by $\upperbound{z}{K,y} - \lowerbound{z}{K,\y}$. It corresponds to an upper bound on the worst-case logit difference between the true class \y and any other class $y$.

\paragraph{Elision of the last layer.}
Bound propagation is not necessary for the last linear layer of the network.
Indeed, we can find an upper bound to the solution of \eqref{eq:verification} that is tighter than proposed by \eqref{eq:upper_bound_verification} by eliding the final linear layer with the specification.
Assuming $h_K(z_{K-1}) = W z_{K-1} + b$, we have:
\begin{equation}
\begin{aligned}
&\underset{\substack{\lbound{z}{K} \leq z_K \leq \ubound{z}{K}\\ z_{K}=h_{K}(z_{K-1})} }{\maximize} \hspace{.5cm} c\T z_K + d \\
&\geq \underset{\lbound{z}{K-1} \leq z_{K-1} \leq \ubound{z}{K-1}}{\maximize} c\T h_K(z_{K-1}) + d \\
&= \underset{\lbound{z}{K-1} \leq z_{K-1} \leq \ubound{z}{K-1}}{\maximize} c\T W z_{K-1} + c\T b + d \\
&= \underset{\lbound{z}{K-1} \leq z_{K-1} \leq \ubound{z}{K-1}}{\maximize} {c'}\T z_{K-1} + d' \\
\end{aligned}
\label{eq:tighter_upper_bound}
\end{equation}
with $c' = W\T c$ and $d' = c\T b + d$, which bypasses the additional relaxation induced by the last linear layer.

\paragraph{Loss.}
In the context of classification under adversarial perturbation, solving the optimization problem~\eqref{eq:upper_bound_verification} for each target class $y \neq \y$ results in a set of worst-case logits where the logit of the true class is equal to its lower bound and the other logits are equal to their upper bound:
\begin{equation}
    \hat{z}_{K,y}(\epsilon) = \left\{
    \begin{aligned}
    & \upperbound{z}{K,y} \quad & & \textrm{if~} y \neq \y \\
    & \lowerbound{z}{K,\y} & & \textrm{otherwise}
    \end{aligned} \right.
\label{eq:worse_logits}
\end{equation}
That is for all $y \neq \y$, we have
\begin{equation}
     (\onehot{y} - \onehot{\y})\T \hat{z}_K(\epsilon) =\!\!\!\! \underset{\lowerbound{z}{K} \leq z_K \leq \upperbound{z}{K}}{\maximize} (\onehot{y} - \onehot{\y})\T z_K
\end{equation}
We can then formulate our training loss as
\begin{equation}
    L = \kappa \underbrace{\ell(z_K, \y)}_{L_\textrm{fit}} + (1 - \kappa) \underbrace{\ell(\hat{z}_K(\epsilon), \y)}_{L_\textrm{spec}}
\label{eq:loss}
\end{equation}
where $\ell$ is the cross-entropy loss and $\kappa$ is a hyperparameter that governs the relative weight of satisfying the specification ($L_\textrm{spec}$) versus fitting the data ($L_\textrm{fit}$).
If $\epsilon = 0$ then $z_K = \hat{z}_K(\epsilon)$, and thus \eqref{eq:loss} becomes equivalent to a standard classification loss.

\paragraph{Training procedure.}
To stabilize the training process and get a good trade-off between nominal and verified accuracy under adversarial perturbation, we create a learning curriculum by scheduling the values of $\kappa$ and $\epsilon$:
\begin{itemize}[topsep=0pt,itemsep=2pt,partopsep=1pt, parsep=1pt]
    \item $\kappa$ controls the relative weight of satisfying the specification versus fitting the data. Hence, we found that starting with $\kappa = 1$ and slowly reducing it throughout training helps get more balanced models with higher nominal accuracy. In practice, we found that using a final value of $\kappa = 1/2$ works well on \mnist, \cifar, \svhn and \imagenet.
    \item More importantly, we found that starting with $\epsilon = 0$ and slowly raising it up to a target perturbation radius $\epsilon_\textrm{train}$ is necessary. We note that $\epsilon_\textrm{train}$ does not need to be equal to the perturbation radius used during testing, using higher values creates robust models that generalize better.
\end{itemize}
Additional details that relate to specific datasets are available in the supplementary material in Appendix~\appendixref{appendix:params}{A}.

% \paragraph{Remark.}
% Equation~\eqref{eq:bound_propagation} highlights that IBP considers hidden units independently.
% There are significant efforts~\cite{raghunathan2018semidefinite} being made on SDP-based verification of neural networks (which allows to take correlations between activations into account).
% However, so far, such approaches are still too slow to be used during training.

\section{Results}
\label{sec:results}

We demonstrate that IBP can train verifiably robust networks and compare its performance to state-of-the-art methods on \mnist, \cifar and \svhn.
Highlights include an improvement of the verified error rate from 3.67\% to 2.23\% on \mnist at $\epsilon = 0.1$, from 19.32\% to 8.05\% on \mnist at $\epsilon = 0.3$, and from 78.22\% to 67.96\% on \cifar at $\epsilon = 8/255$.
We also show that IBP can scale to larger networks by training a model on downscaled \imagenet that reaches a non-vacuous verified error rate of $93.87\%$ at $\epsilon = 1/255$.
Finally, Section~\ref{sec:tightness} illustrates how training with the loss function and curriculum from Section~\ref{sec:ibp} allows the training process to adapt the model to ensure that the bound computed by IBP is tight.

Unless stated otherwise, we compute the empirical adversarial accuracy (or error rate) on the test set using 200 untargeted PGD steps and 10 random restarts.
As the verified error rate computed for a network varies greatly with the verification method, we calculate it using an exact solver.
Several previous works have shown that training a network with a loss function derived from a specific verification procedure renders the network amenable to verification using that specific procedure only \citep{wong2018provable, raghunathan2018certified, dvijotham2018training}. 
In order to circumvent this issue and present a fair comparison, we use a complete verification algorithm based on solving a MIP -- such an algorithm is expensive as it performs a brute force enumeration in the worst case. 
However, in practice, we find that commercial MIP solvers like Gurobi can handle verification problems from moderately sized networks. 
In particular, we use the MIP formulation from \citet{tjeng2017verifying}.
For each example of the test set, a MIP is solved using Gurobi with a timeout of 10 minutes.
Upon timeout, we fallback to solving a relaxation of the verification problem with a LP~\citep{ehlers2017formal} using Gurobi again.
When both approaches fail to provide a solution within the imparted time, we count the example as attackable.
Thus, the verified error rate reported may be over-estimating the exact verified error rate.\footnote{As an example, for the small model trained using \citeauthor{wong2018scaling}, there are 3 timeouts at $\epsilon = 0.1$, 18 timeouts at $\epsilon = 0.2$ and 58 timeouts at $\epsilon = 0.3$ for the 10K examples of the \mnist test set. These timeouts would amount to a maximal over-estimation of 0.03\%, 0.18\% and 0.58\% in verified error rate, respectively.}
We always report results with respect to the complete test set of 10K images for both \mnist and \cifar, and 26K images for \svhn.
For downscaled \imagenet, we report results on the validation set of 10K images.

\subsection{\mnist, \cifar and \svhn}

\begin{table}[t]
\begin{center}
\begin{footnotesize}
\begin{tabular}{ccc}
\hline
small & medium & large \\
\hline
\textsc{Conv} $16$ $4\!\!\times\!\!4\!\!+\!\!2$
& \textsc{Conv} $32$ $3\!\!\times\!\!3\!\!+\!\!1$
& \textsc{Conv} $64$ $3\!\!\times\!\!3\!\!+\!\!1$ \\
\textsc{Conv} $32$ $4\!\!\times\!\!4\!\!+\!\!1$
& \textsc{Conv} $32$ $4\!\!\times\!\!4\!\!+\!\!2$
& \textsc{Conv} $64$ $3\!\!\times\!\!3\!\!+\!\!1$ \\
\textsc{FC} $100$
& \textsc{Conv} $64$ $3\!\!\times\!\!3\!\!+\!\!1$
& \textsc{Conv} $128$ $3\!\!\times\!\!3\!\!+\!\!2$ \\
& \textsc{Conv} $64$ $4\!\!\times\!\!4\!\!+\!\!2$
& \textsc{Conv} $128$ $3\!\!\times\!\!3\!\!+\!\!1$ \\
& \textsc{FC} $512$
& \textsc{Conv} $128$ $3\!\!\times\!\!3\!\!+\!\!1$ \\
& \textsc{FC} $512$
& \textsc{FC} $512$ \\
\hline
\multicolumn{1}{l}{\textbf{\# hidden:} ~8.3K} & 47K & 230K \\
\multicolumn{1}{l}{\textbf{\# params:} 471K} & 1.2M & 17M \\
\hline
\end{tabular}
\end{footnotesize}
\end{center}
\caption{
Architecture of the three models used on \mnist, \cifar and \svhn.
All layers are followed by \textsc{ReLU} activations.
The last fully connected layer is omitted.
``\textsc{Conv} $k$ $w\!\times\!h\!+\!s$'' corresponds to a 2D convolutional layer with $k$ filters of size $w\!\times\!h$ using a stride of $s$ in both dimensions.
``\textsc{FC} $n$'' corresponds to a fully connected layer with $n$ outputs.
The last two rows are the number of hidden units (counting activation units only) and the number of parameters when training on \cifar.
}
\label{table:architecture}
\end{table}

\begin{figure*}[t]
\centering
\begin{subfigure}{.32\textwidth}
\includegraphics[width=\linewidth]{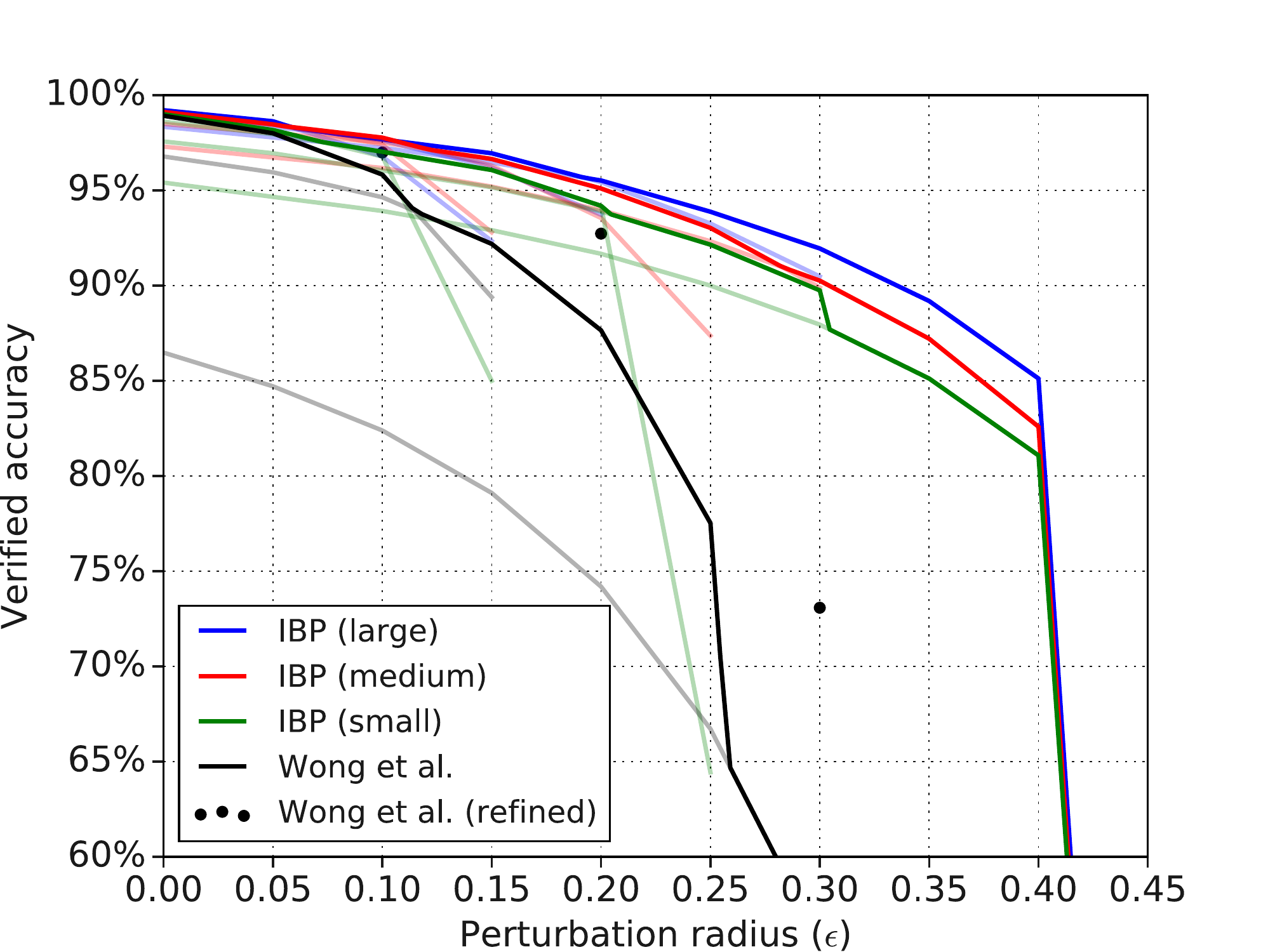}
\caption{\label{fig:verified_accuracy}}
\end{subfigure}
\begin{subfigure}{.32\textwidth}
\includegraphics[width=\linewidth]{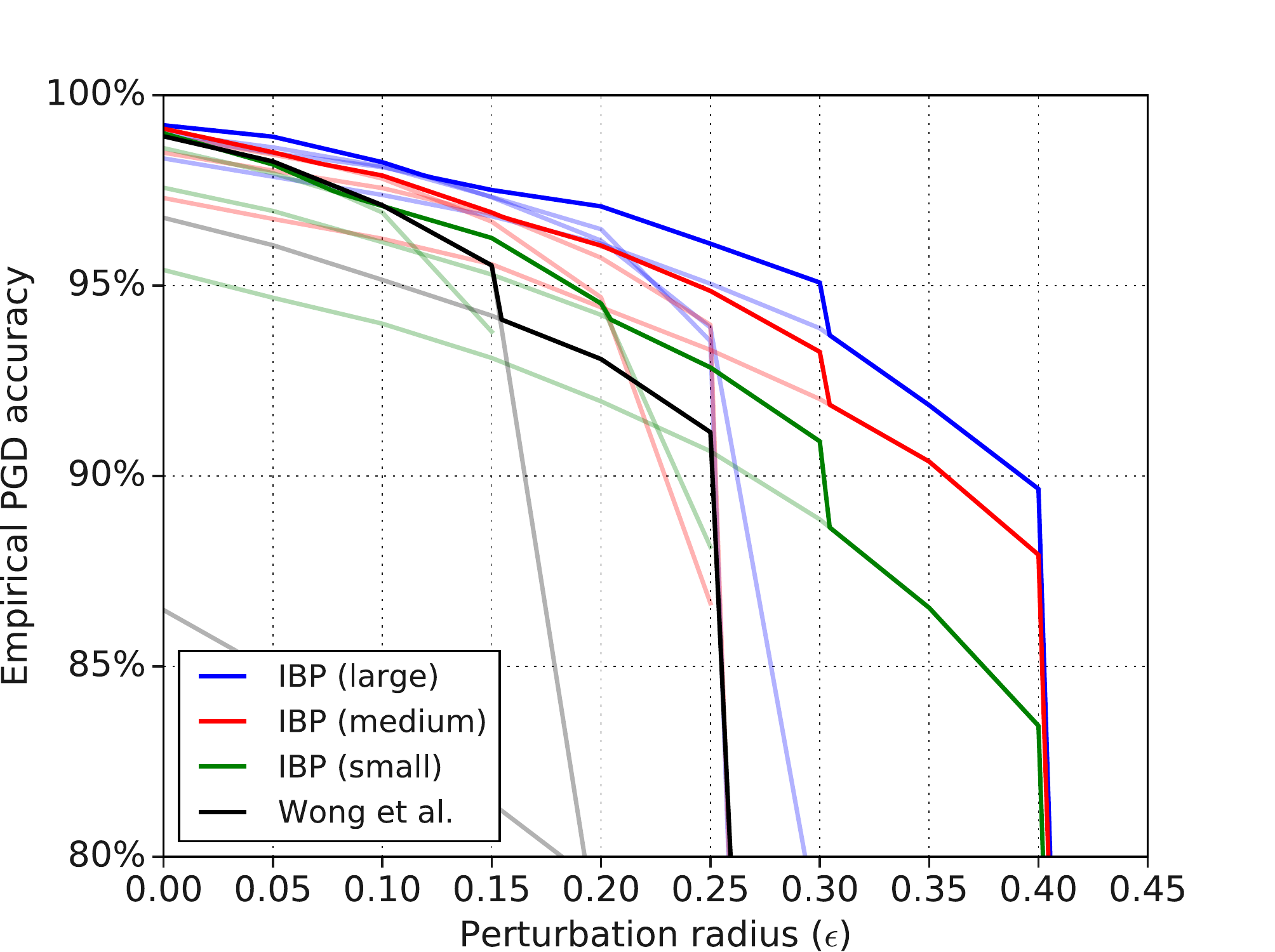}
\caption{\label{fig:attacked_accuracy_wong}}
\end{subfigure}
\begin{subfigure}{.32\textwidth}
\includegraphics[width=\linewidth]{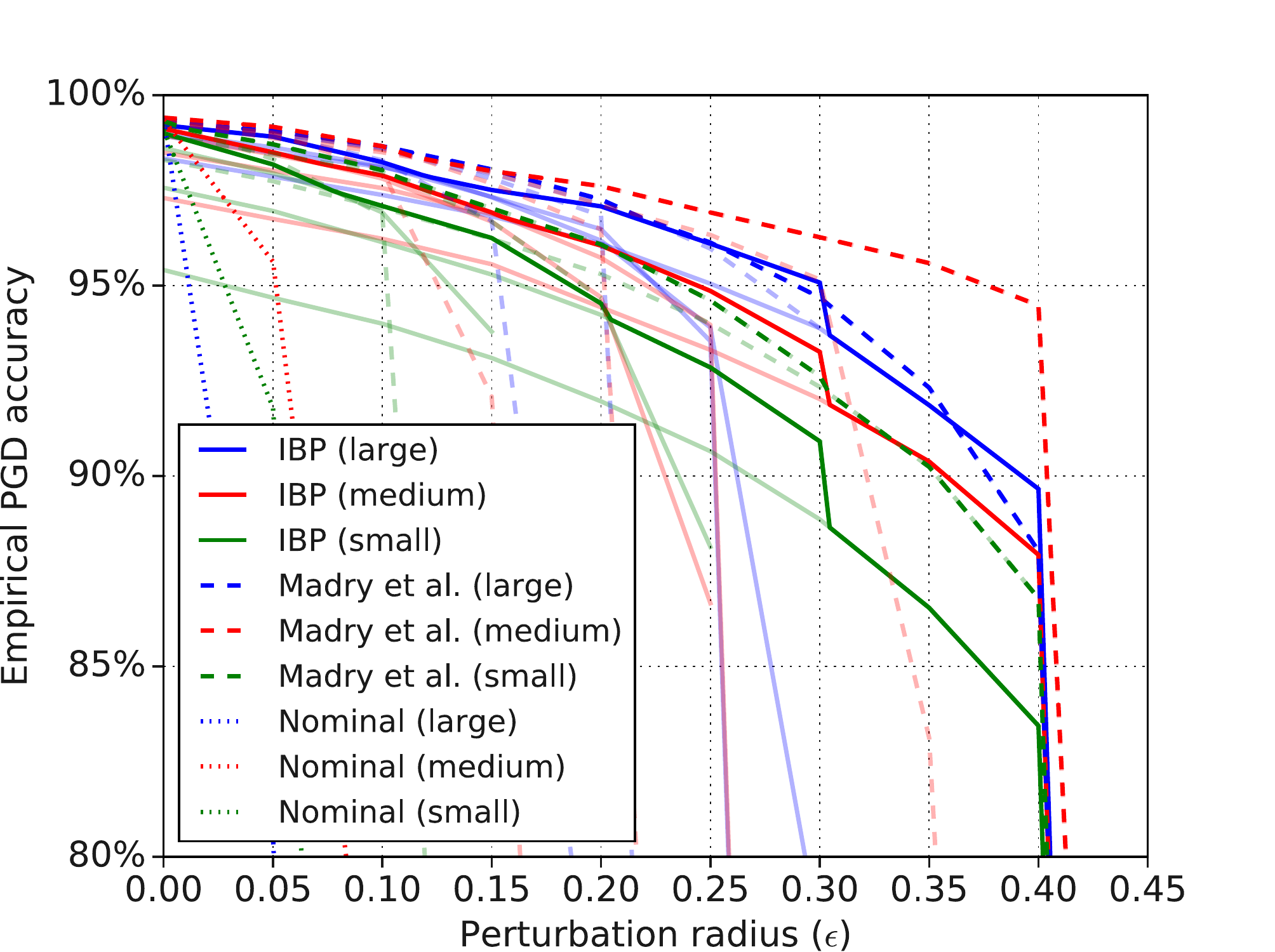}
\caption{\label{fig:attacked_accuracy_madry}}
\end{subfigure}
\caption{
Accuracy against different adversarial perturbations: (\subref{fig:verified_accuracy}) shows the verified/provable worst-case accuracy compared to \citeauthor{wong2018scaling}, (\subref{fig:attacked_accuracy_wong}) shows the empirical adversarial accuracy computed by running PGD  compared to \citeauthor{wong2018scaling}, and (\subref{fig:attacked_accuracy_madry}) shows the empirical adversarial accuracy computed by running PGD compared to \citeauthor{madry2017towards}. Faded lines show individual models of a given size (i.e., small, medium and large) trained with $\epsilontrain = \{ 0.1, 0.2, 0.3, 0.4 \}$, while bold lines show the best accuracy across across all \epsilontrain values for each model size. In~(\subref{fig:verified_accuracy}), for \citeauthor{wong2018scaling}, the dots correspond to exact bounds computed using a MIP solver, while the black bold line corresponds to a lower bound computed using \citep{wong2018scaling} without random projections.
}
\label{fig:results_wong_madry}
\end{figure*}

We compare IBP to three alternative approaches:
the nominal method, which corresponds to standard training with cross-entropy loss;
adversarial training, following \citet{madry2017towards}, which generates adversarial examples on the fly during training;
and \citet{wong2018scaling}, which trains models that are provably robust.
We train three different model architectures for each of the four methods (see Table~\ref{table:architecture}).
The first two models (i.e., \textbf{small} and \textbf{medium}) are equivalent to the small and large models in~\citet{wong2018scaling}.
\footnote{We do not train our large model with \citeauthor{wong2018scaling} as we could not scale this method beyond the medium sized model.}
The third model (i.e., \textbf{large}) is significantly larger (in terms of number of hidden units) than any other verified model presented in the literature.
On \mnist, for each model and each method, we trained models that are robust to a wide range of perturbation radii by setting \epsilontrain to $0.1, 0.2, 0.3$ or $0.4$.
During testing, we test each of these 12 models against $\epsilon \in [0, 0.45]$.
On \cifar, we train the same models and methods with $\epsilon_\textrm{train} \in \{2/255, 8/255\}$ and test on the same $\epsilon = \epsilontrain$ value.
On \svhn we used $\epsilontrain = 0.01$ and only test on $\epsilon = \epsilontrain$.

Figures~\ref{fig:verified_accuracy} and \subref{fig:attacked_accuracy_wong} compare IBP to \citeauthor{wong2018scaling} on \mnist for all perturbation radii between 0 and 0.45 across all models.
Remember that we trained each model architecture against many \epsilontrain values.
The bold lines show for each model architecture, the model trained with the perturbation radius \epsilontrain that performed best for a given $\epsilon$ (i.e., x-axis)
The faded lines show all individual models.
Across the full spectrum, IBP achieves good accuracy under PGD attacks and higher provable accuracy (computed by an exact verifier).
We observe that while \citeauthor{wong2018scaling} is competitive at small perturbation radii (when both $\epsilon$ and \epsilontrain are small), it quickly degrades as the perturbation radius increases (when \epsilontrain is large).
For completeness, Figure~\ref{fig:attacked_accuracy_madry} also compares IBP to \citeauthor{madry2017towards} with respect to the empirical accuracy against PGD attacks of varying intensities.
We observe that IBP tends to be slightly worse than \citeauthor{madry2017towards} for similar network sizes -- except for the large model where \citeauthor{madry2017towards} is likely overfitting (as it performs worse than the medium-sized model).

Table~\ref{table:full_results} provides additional results and also includes results from the literature.
The test error corresponds to the test set error rate when there is no adversarial perturbation.
For models that we trained ourselves, the PGD error rate is calculated using 200 iterations of PGD and 10 random restarts.
The verified bound on the error rate is obtained using the MIP/LP cascade described earlier.
A dash is used to indicate when we could not verify models beyond trivial bounds within the imparted time limits.
For such cases, it is useful to consider the PGD error rate as a lower bound on the verified error rate.
All methods use the same model architectures (except results from the literature).
For clarity, we do not report the results for all $\epsilon_\textrm{train}$ values and all model architectures (Table~\appendixref{table:additional_results}{6} in the appendix reports additional results).
Figure~\ref{fig:results_wong_madry} already shows the effect of $\epsilon_\textrm{train}$ and model size in a condensed form.
Compared to \citeauthor{wong2018scaling}, IBP achieves lower error rates under normal and adversarial conditions, as well as better verifiable bounds, setting the state-of-the-art in verified robustness to \linf-bounded adversarial attacks on most pairs of dataset and perturbation radius.
Additionally, IBP remains competitive against \citeauthor{madry2017towards} by achieving a lower PGD error rate on \cifar with $\epsilon = 8/255$ (albeit at the cost of an increased nominal error rate).\footnote{This result only holds for our constrained set of network sizes. The best known empirical adversarial error rate for \cifar at $\epsilon = 8/255$ using \citeauthor{madry2017towards} is 52.96\% when using 20 PGD steps and no restarts. As a comparison, our large model on \cifar achieves an empirical adversarial error rate of 60.1\% when using 20 PGD steps and no restarts.}
\cifar with $\epsilon = 2/255$ is the only combination where IBP is worse than \citet{wong2018scaling}.
From our experience, the method from \citeauthor{wong2018scaling} is more effective when the perturbation radius is small (as visible on Figure~\ref{fig:verified_accuracy}), thus giving a marginally better feedback when training on \cifar at $\epsilon = 2/255$.
Additional results are available in Appendix~\appendixref{appendix:additional_results}{D}.
Appendix~\appendixref{appendix:ablation}{C} also details an ablative study that demonstrates that {\it(i)} using cross-entropy (rather than a hinge or softplus loss on each specification) improves verified accuracy across all datasets and model sizes, that {\it(ii)} eliding the last linear layer also provides a small but consistent improvement (especially for models with limited capacity), and that {\it(iii)} the schedule on $\epsilon$ is necessary.

Finally, we note that when training the small network on \mnist with a Titan Xp GPU (where standard training takes 1.5 seconds per epoch), IBP only takes 3.5 seconds per epoch compared to 8.5 seconds for \citeauthor{madry2017towards} and 2 minutes for \citeauthor{wong2018scaling} (using random projection of 50 dimensions).
Indeed, as detailed in Section~\ref{sec:nbp} (under the paragraph interval bound propagation), IBP creates only two additional passes through the network compared to \citeauthor{madry2017towards} for which we used seven PGD steps.

\begin{table}[t]
\begin{center}
\begin{footnotesize}
% dataset, epsilon, model, nominal error rate, lower bound, upper bound.
\begin{tabular}{l|lrrr}
    \hline
    {\bf $\epsilon$} & {\bf Method} & {\bf Test error} & {\bf PGD} & {\bf Verified} \\
    \hline
    \multirow{3}{*}{$1/255$}
      & Nominal & \textbf{48.84\%} & 100.00\% & -- \\
      & \citeauthor{madry2017towards} & 51.52\% & \textbf{70.03\%} & -- \\
      & \cellcolor{Highlight} IBP & \cellcolor{Highlight} 84.04\% & \cellcolor{Highlight} 90.88\% & \cellcolor{Highlight} \textbf{93.87\%} \\
    \hline
\end{tabular}
\end{footnotesize}
\end{center}
\caption{
{\bf Downscaled \imagenet results.}
Comparison of the nominal test error (under no perturbation), empirical PGD error rate, and verified bound on the error rate.
The verified error rate is computed using IBP bounds only, as running a complete solver is too slow for this model.
}
\label{table:imagenet_results}
\end{table}

\begin{figure*}[t]
\centering
\begin{subfigure}{0.31\textwidth}
\includegraphics[width=\linewidth]{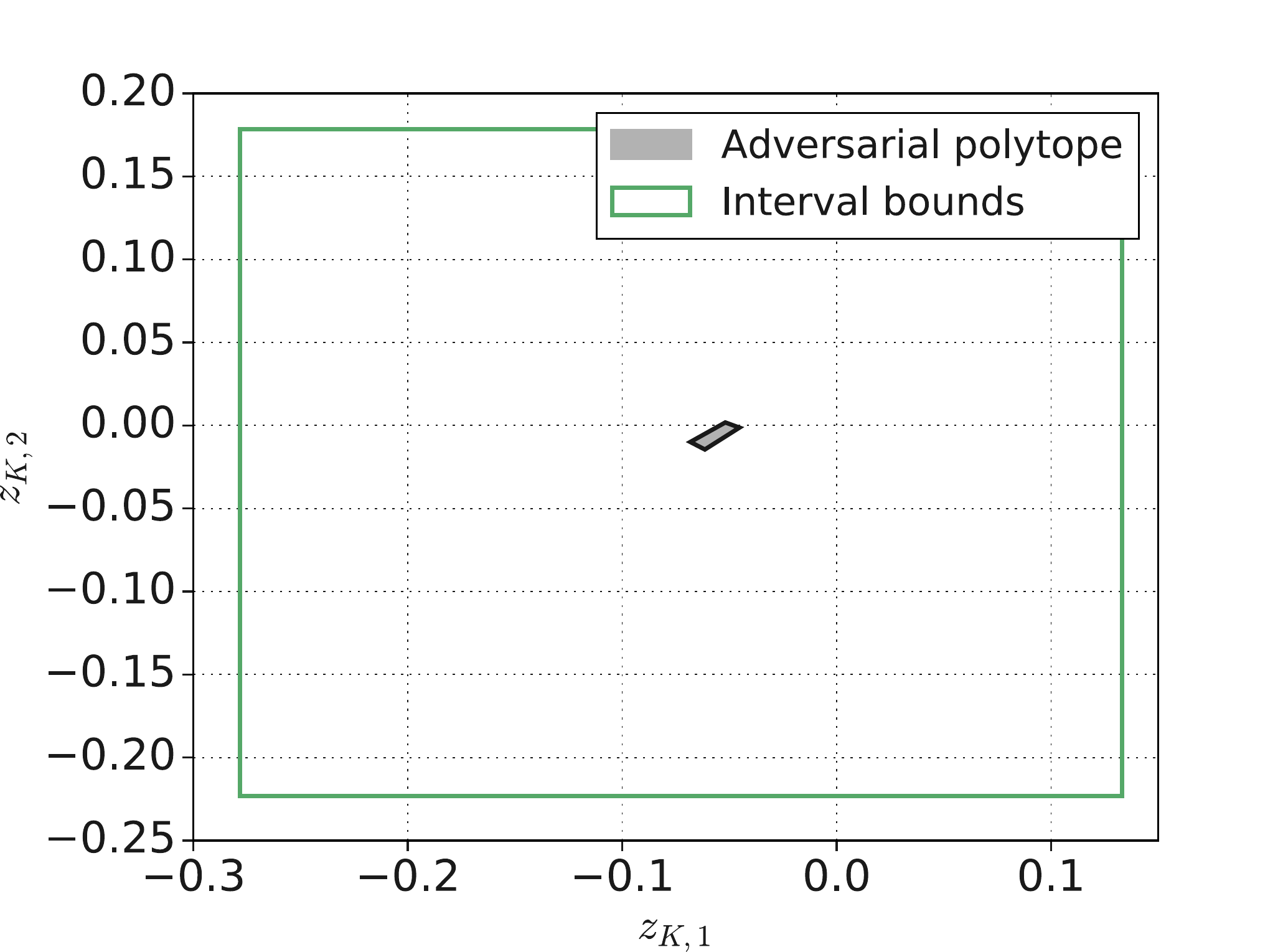}
\caption{Training step 0}
\end{subfigure}
\begin{subfigure}{0.31\textwidth}
\includegraphics[width=\linewidth]{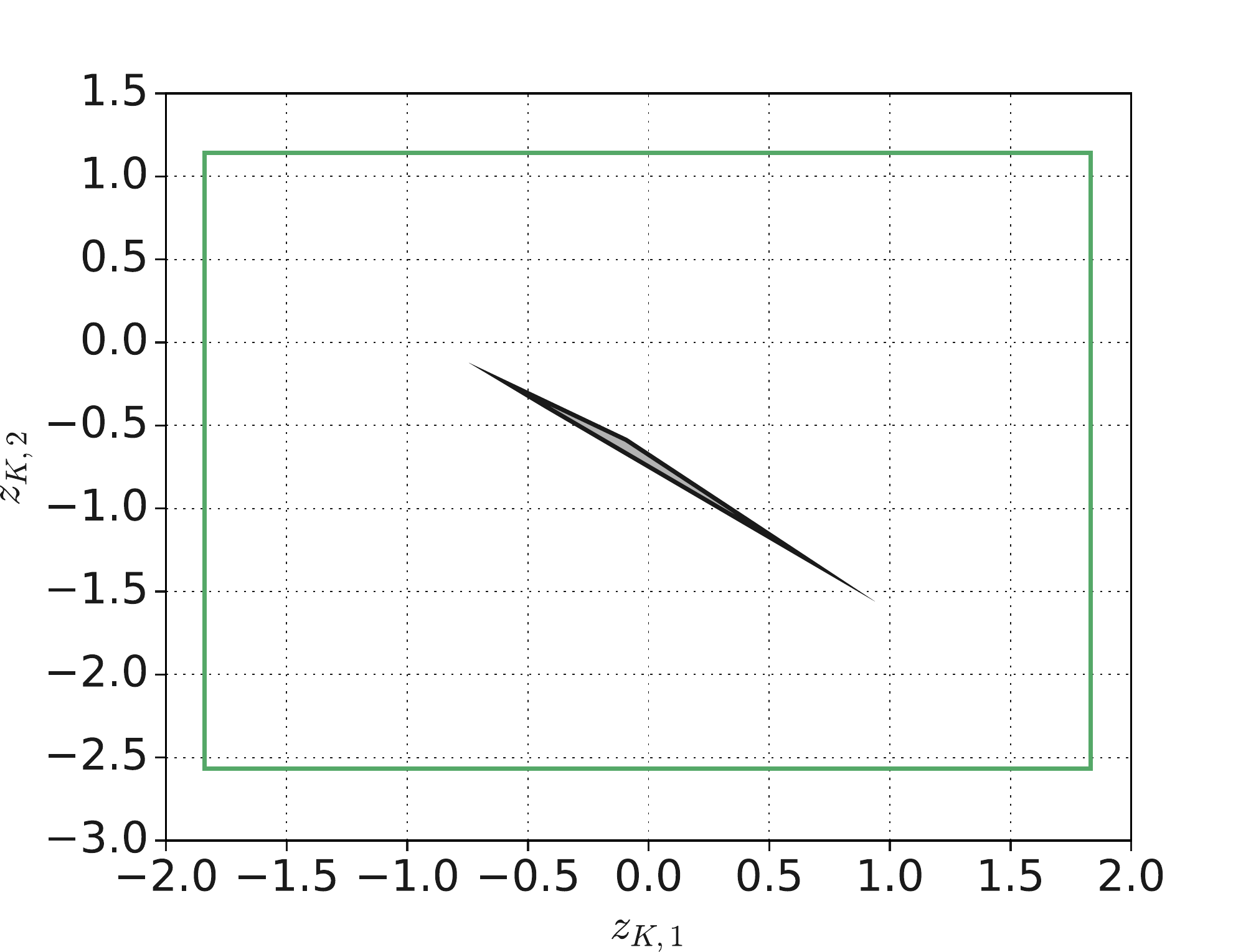}
\caption{Training step 200}
\end{subfigure}
\begin{subfigure}{0.31\textwidth}
\includegraphics[width=\linewidth]{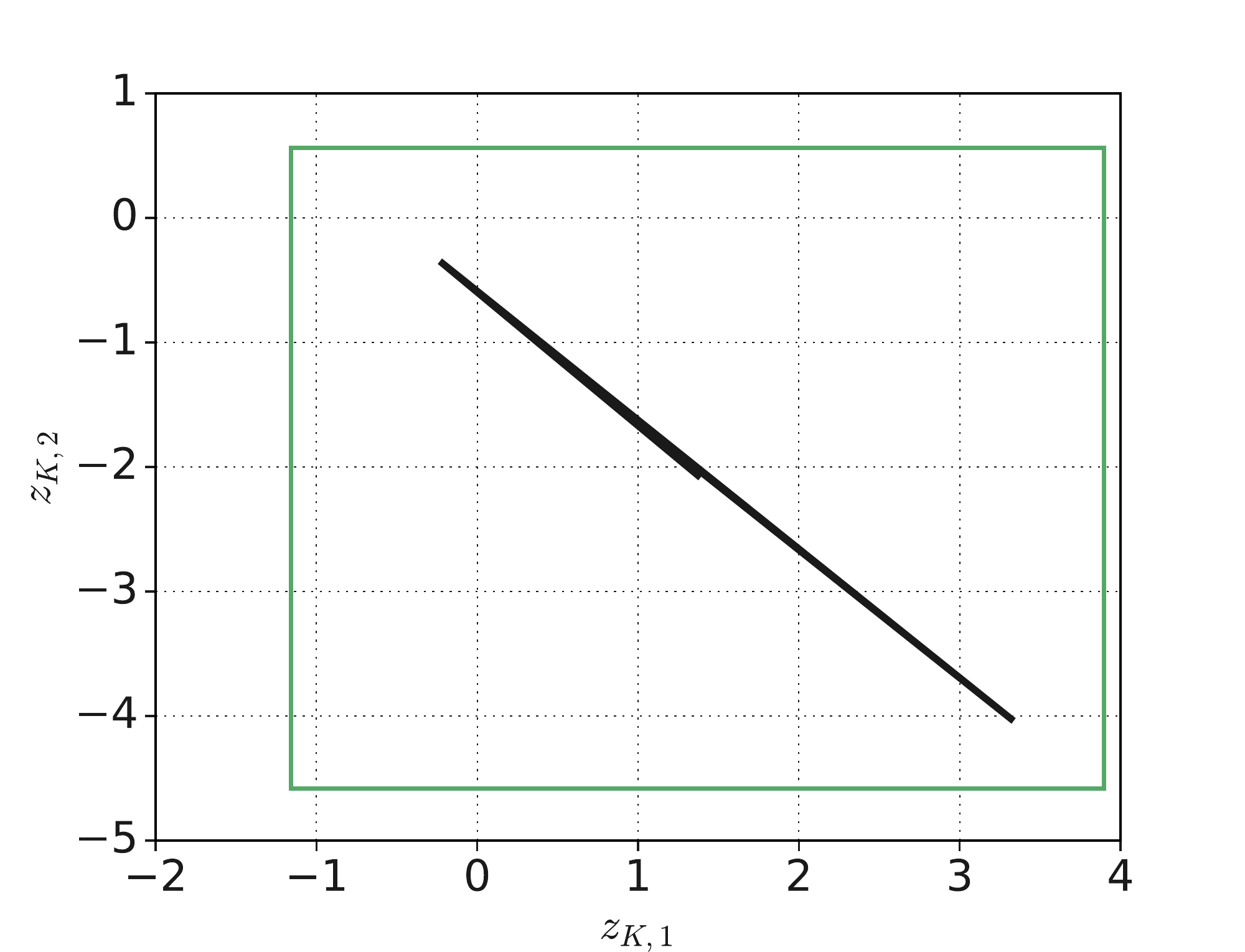}
\caption{Training step 800}
\end{subfigure}
\caption{Evolution of the adversarial polytope (in gray) around the same input during training. The outer approximation computed using IBP is shown in green.}
\label{fig:polytope}
\end{figure*}

\subsection{Downscaled \imagenet}

This section demonstrates the scalability of IBP by training, to the best of our knowledge, the first model with non-vacuous verifiable bounds on \imagenet.
We train a \wideresnet-10-10 with 8M parameters and 1.8M hidden units, almost an order of magnitude greater than the number of hidden units in our large network.
The results in Table~\ref{table:imagenet_results} are obtained through standard non-robust training, adversarial training, and robust training using IBP on downscaled images (i.e., $64 \times 64$).
We use all 1000 classes and measure robustness (either empirical or verifiably) using the same one-vs-all scheme used for \mnist, \cifar and \svhn.
Additional details are available in the supplementary material in Appendix~\appendixref{appendix:params}{A}.
We realize that these results are pale in comparison to the nominal accuracy obtained by larger non-robust models (i.e., \citet{real2018regularized} achieving 16.1\% top-1 error rate).
However, we emphasize that no other work has formally demonstrated robustness to norm-bounded perturbation on \imagenet, even for small perturbations like $\epsilon = 1/255$.

\subsection{Tightness}
\label{sec:tightness}

Figure~\ref{fig:polytope} shows the evolution of an adversarial polytope and its outer approximation during training.
In this setup (similar to \citealp{wong2018provable}), we train a 2-100-100-100-2 network composed of fully-connected layers with ReLU activations on a toy two-dimensional dataset.
This dataset consists of 13 randomly drawn 2-dimensional points in $[0, 1]^2$, five of which are from the positive class.
The \linf distance between each pair of points is at least 0.08, which corresponds to the $\epsilon$ and \epsilontrain values used during testing and training, respectively.
The adversarial polytope at the last layer (shown in gray) is computed by densely sampling inputs within an \linf-norm bounded ball around a nominal input (corresponding to one of the positive training examples).
The outer bounds (in green) correspond to the interval bounds at the last layer computed using \eqref{eq:bound_propagation}.
We observe that, while initially the bounds are very loose, they do become tighter as training progresses.

To judge the tightness of IBP quantitatively, we compare the final verified error rate obtained using the MIP/LP cascade describe earlier with the upper bound estimates from IBP only.
Table~\ref{table:tightness} shows the differences.
We observe that IBP itself is a good estimate of the verified error rate and provides estimates that are competitive with more sophisticated solvers (when models are trained using IBP).
While intuitive, it is surprising to see that the IBP bounds are so close to the MIP bounds.
This highlights that verification becomes easier when models are trained to be verifiable as a simple method like IBP can verify a large proportion of the MIP verified samples.
This phenomenon was already observed by \citet{dvijotham2018training} and \citet{xiao2018training} and explains why some methods cannot be verified beyond trivial bounds within a reasonable computational budget.

\begin{table}[t]
\begin{center}
\begin{footnotesize}
\begin{tabular}{ll|rr}
    \hline
    {\bf Dataset} & {\bf Epsilon} & {\bf IBP bound} & {\bf MIP bound} \\
    \hline
    \multirow{4}{*}{\mnist}
    & $\epsilon = 0.1$ & 2.92\% & 2.23\% \\
    & $\epsilon = 0.2$ & 4.53\% & 4.48\% \\
    & $\epsilon = 0.3$ & 8.21\% & 8.05\% \\
    & $\epsilon = 0.4$ & 15.01\% & 14.88\% \\
    \hline
    \multirow{2}{*}{\cifar}
    & $\epsilon = 2/255$ & 55.88\% & 49.98\% \\
    & $\epsilon = 8/255$ & 68.44\% & 67.96\% \\
    \hline
    \multirow{1}{*}{\svhn}
    & $\epsilon = 0.01$ & 39.35\% & 37.60\% \\
    \hline
\end{tabular}
\end{footnotesize}
\end{center}
\caption{
{\bf Tightness of IBP verified bounds on the error rate.}
This table compares the verified bound on the error rate obtained using the MIP/LP cascade with the estimates from IBP only (obtained using the worst-case logits from \eqref{eq:worse_logits}).
The models are the ones reported in Table~\ref{table:full_results}.
}
\label{table:tightness}
\end{table}

\section{Conclusion}

We have presented an approach for training verifiable models and provided strong baseline results for \mnist, \cifar, \svhn and downscaled \imagenet.
Our experiments have shown that the proposed approach outperforms competing techniques in terms of verified bounds on adversarial error rates in image classification problems, while also training faster.
In the future, we hope that these results can serve as a useful baseline.
We believe that this is an important step towards the vision of specification-driven ML.

% \section*{Reproducibility}
% The code for training verifiably robust neural networks using IBP is available at {\footnotesize\url{https://github.com/deepmind/interval-bound-propagation}}.

\begin{table*}[tb]
\begin{center}
\begin{footnotesize}
% dataset, epsilon, model, nominal error rate, lower bound, upper bound.
\begin{tabular}{ll|lrrr}
    \hline
    {\bf Dataset} & {\bf Epsilon} & {\bf Method} & {\bf Test error} & {\bf PGD} & {\bf Verified} \\
    \hline
    \multirow{8}{*}{\mnist} & \multirow{8}{*}{$\epsilon = 0.1$}
      & Nominal & 0.65\% & 27.72\% & -- \\
    & & \citeauthor{madry2017towards} ($\epsilon_\textrm{train} = 0.2$) & \textbf{0.59\%} & \textbf{1.34\%} & -- \\
    & & \citeauthor{wong2018scaling} ($\epsilon_\textrm{train} = 0.1$) & 1.08\% & 2.89\% & 3.01\% \\
    & & \cellcolor{Highlight} IBP ($\epsilon_\textrm{train} = 0.2$) & \cellcolor{Highlight} 1.06\% & \cellcolor{Highlight} 2.11\% & \cellcolor{Highlight} \textbf{2.23\%} \\
    \cline{3-6}
    & & \multicolumn{4}{l}{{\bf Reported in literature}*} \\
    & & \hspace{.5cm}\citet{xiao2018training}** & 1.05\% & 3.42\% & 4.40\% \\
    & & \hspace{.5cm}\citet{wong2018scaling} & 1.08\% & -- & 3.67\% \\
    & & \hspace{.5cm}\citet{dvijotham2018training} & 1.20\% & 2.87\% & 4.44\% \\
    % & & \hspace{.5cm}\citet{mirman18b}*** & 1.0\% & 2.4\% & 3.4\% \\
    
    \hline
    \multirow{6}{*}{\mnist} & \multirow{6}{*}{$\epsilon = 0.2$}
      & Nominal & \textbf{0.65\%} & 99.57\% & -- \\
    & & \citeauthor{madry2017towards} ($\epsilon_\textrm{train} = 0.4$) & 0.70\% & \textbf{2.39\%} & -- \\
    & & \citeauthor{wong2018scaling} ($\epsilon_\textrm{train} = 0.2$) & 3.22\% & 6.93\% & 7.27\% \\
    & & \cellcolor{Highlight} IBP ($\epsilon_\textrm{train} = 0.4$) & \cellcolor{Highlight} 1.66\% & \cellcolor{Highlight} 3.90\% & \cellcolor{Highlight} \textbf{4.48\%} \\
    \cline{3-6}
    & & \multicolumn{4}{l}{\bf Reported in literature} \\
    & & \hspace{.5cm}\citet{xiao2018training} & 1.90\% & 6.86\% & 10.21\% \\
    
    \hline
    \multirow{8}{*}{\mnist} & \multirow{8}{*}{$\epsilon = 0.3$}
      & Nominal & \textbf{0.65\%} & 99.63\% & -- \\
    & & \citeauthor{madry2017towards} ($\epsilon_\textrm{train} = 0.4$) & 0.70\% & \textbf{3.73\%} & -- \\
    & & \citeauthor{wong2018scaling} ($\epsilon_\textrm{train} = 0.3$) & 13.52\% & 26.16\% & 26.92\% \\
    & & \cellcolor{Highlight} IBP ($\epsilon_\textrm{train} = 0.4$) & \cellcolor{Highlight} 1.66\% & \cellcolor{Highlight} 6.12\% & \cellcolor{Highlight} \textbf{8.05\%} \\
    \cline{3-6}
    & & \multicolumn{4}{l}{\bf Reported in literature} \\
    & & \hspace{.5cm}\citet{madry2017towards} & 1.20\% & 6.96\% & -- \\
    & & \hspace{.5cm}\citet{xiao2018training} & 2.67\% & 7.95\% & 19.32\% \\
    & & \hspace{.5cm}\citet{wong2018scaling} & 14.87\% & -- & 43.10\% \\
    
    \hline
    \multirow{3}{*}{\mnist} & \multirow{3}{*}{$\epsilon = 0.4$}
      & Nominal & \textbf{0.65\%} & 99.64\% & -- \\
    & & \citeauthor{madry2017towards} ($\epsilon_\textrm{train} = 0.4$) & 0.70\% & \textbf{5.52\%} & -- \\
    & & \cellcolor{Highlight} IBP ($\epsilon_\textrm{train} = 0.4$) & \cellcolor{Highlight} 1.66\% & \cellcolor{Highlight} 10.34\% & \cellcolor{Highlight} \textbf{14.88\%} \\
    
    \hline
    \multirow{7}{*}{\cifar} & \multirow{7}{*}{$\epsilon = 2/255$}
      & Nominal & 16.66\% & 87.24\% & -- \\
    & & \citeauthor{madry2017towards} ($\epsilon_\textrm{train} = 2/255$) & \textbf{15.54\%} & \textbf{42.01\%} & -- \\
    & & \citeauthor{wong2018scaling} ($\epsilon_\textrm{train} = 2/255$) & 36.01\% & 45.11\% & 49.96\% \\
    & & \cellcolor{Highlight} IBP ($\epsilon_\textrm{train} = 2/255$) & \cellcolor{Highlight} 29.84\% & \cellcolor{Highlight} 45.09\% & \cellcolor{Highlight} 49.98\% \\
    \cline{3-6}
    & & \multicolumn{4}{l}{\bf Reported in literature} \\
    & & \hspace{.5cm}\citet{xiao2018training} & 38.88\% & 50.08\% & 54.07\% \\
    & & \hspace{.5cm}\citet{wong2018scaling} & 31.72\% & -- & \textbf{46.11\%} \\
    % & & \hspace{.5cm}\citet{mirman18b}*** & 38.0\% & 45.4\% & 47.8\% \\
    
    \hline
    \multirow{9}{*}{\cifar} & \multirow{9}{*}{$\epsilon = 8/255$}
      & Nominal & 16.66\% & 100.00\% & 100.00\% \\
    & & \citeauthor{madry2017towards} ($\epsilon_\textrm{train} = 8/255$) & 20.33\% & 75.95\% & -- \\
    & & \citeauthor{wong2018scaling} ($\epsilon_\textrm{train} = 8/255$) & 71.03\% & 78.14\% & 79.21\% \\
    & & \cellcolor{Highlight} IBP ($\epsilon_\textrm{train} = 8/255$) & \cellcolor{Highlight} 50.51\% & \cellcolor{Highlight} 65.23\% & \cellcolor{Highlight} \textbf{67.96\%} \\
    \cline{3-6}
    & & \multicolumn{4}{l}{\bf Reported in literature} \\
    & & \hspace{.5cm}\citet{madry2017towards} & \textbf{12.70\%} & \textbf{52.96\%} & -- \\
    & & \hspace{.5cm}\citet{xiao2018training} & 59.55\% & 73.22\% & 79.73\% \\
    & & \hspace{.5cm}\citet{wong2018scaling} & 71.33\% & -- & 78.22\% \\
    & & \hspace{.5cm}\citet{dvijotham2018training}*** & 51.36\% & 67.28\% & 73.33\% \\
    
    \hline
    \multirow{6}{*}{\svhn} & \multirow{6}{*}{$\epsilon = 0.01$}
      & Nominal & 5.13\% & 94.14\% & -- \\
    & & \citeauthor{madry2017towards} ($\epsilon_\textrm{train} = 0.01$) & 6.18\% & \textbf{29.06\%} & -- \\
    & & \citeauthor{wong2018scaling} ($\epsilon_\textrm{train} = 0.01$) & 18.10\% & 32.41\% & 37.96\% \\
    & & \cellcolor{Highlight} IBP ($\epsilon_\textrm{train} = 0.01$) & \cellcolor{Highlight} 14.82\% & \cellcolor{Highlight} 32.46\% & \cellcolor{Highlight} \textbf{37.60\%} \\
    \cline{3-6}
    & & \multicolumn{4}{l}{\bf Reported in literature} \\
    & & \hspace{.5cm}\citet{wong2018provable} & 20.38\% & 33.74\% & 40.67\% \\
    & & \hspace{.5cm}\citet{dvijotham2018training} & 16.59\% & 33.14\% & \textbf{37.56\%} \\
    
    \hline
\end{tabular}
\end{footnotesize}
\end{center}
\caption{
{\bf Comparison with the state-of-the-art.}
Comparison of the nominal test error (no adversarial perturbation), error rate under PGD attacks, and verified bound on the error rate.
The PGD error rate is calculated using 200 iterations of PGD and 10 random restarts.
Dashes ``--'' indicate that we were unable to verify these networks beyond the trivial 100\% error rate bound within the imparted time limit; for such cases we know that the verified error rate must be at least as large as the PGD error rate.
For the models we trained ourselves, we indicate the \epsilontrain that lead to the lowest verified (or -- when not available -- empirical) error rate.
Results from \citet{mirman2018differentiable} are only included in Appendix~\appendixref{appendix:additional_results}{D} as, due to differences in image normalization, different $\epsilon$ were used; we confirmed with the authors that our IBP results are significantly better.\\
* Results reported from the literature may use different network architectures.
Their empirical PGD error rate may have been computed with a different number of PGD steps and a different number of restarts (when possible we chose the closest setting to ours).
Except for the results from \citet{xiao2018training}, the reported verified bound on the error rate is not computed with an exact solver and may be over-estimated.\\
** For this model, \citet{xiao2018training} only provides estimates computed from 1000 samples (rather than the full 10K images).\\
*** \citet{dvijotham2018training} use a slightly smaller $\epsilon = 0.03 = 7.65/255$ for \cifar.
}
\label{table:full_results}
\end{table*}
\afterpage{\clearpage}

\newpage
\clearpage
\setlength{\bibsep}{0.0pt}
{\small
\bibliographystyle{IEEEtranN}
\bibliography{bibliography}
}

\ifx\cutpaper\undefined
\newpage
\clearpage

\onecolumn
\appendix

\begin{center}
  {\Large \bf Scalable Verified Training for Provably Robust Image Classification (Supplementary Material)}
\end{center}
\vspace{-.5cm}

\section{Training parameters}
\label{appendix:params}

For IBP, across all datasets, the networks were trained using the Adam~\citep{kingma2014adam} algorithm with an initial learning rate of $10^{-3}$.
We linearly ramp-down the value of $\kappa$ between 1 and $\kappa_\textrm{final}$ after a fixed warm-up period ($\kappa_\textrm{final}$ is set to both 0 or 0.5 and the best result is used).
Simultaneously, we lineary ramp-up the value of $\epsilon$ between 0 and \epsilontrain (for \cifar and \svhn, we use a value of \epsilontrain that is 10\% higher than the desired robustness radius).
\mnist is trained on a single Nvidia V100 GPU.
\cifar, \svhn and \imagenet are trained on 32 tensor processing units (TPU) \cite{abadi2016tensorflow} with 2 workers with 16 TPU cores each.

\begin{itemize}[topsep=0pt,itemsep=2pt,partopsep=1pt, parsep=1pt]
    \item For \mnist, we train on a single Nvidia V100 GPU for 100 epochs with batch sizes of 100.
          The total number of training steps is 60K.
          We decay the learning rate by 10$\times$ at steps 15K and 25K.
          We use warm-up and ramp-up durations of 2K and 10K steps, respectively.
          We do not use any data augmentation techniques and use full $28 \times 28$ images without any normalization.
    \item \cifar, we train for 3200 epochs with batch sizes of 1600 (training for 350 epochs with batch sizes of 50 reaches 71.70\% verified error rate when $\epsilon = 8/255$).
          The total number of training steps is 100K.
          We decay the learning rate by 10$\times$ at steps 60K and 90K.
          We use warm-up and ramp-up durations of 5K and 50K steps, respectively.
          During training, we add random translations and flips, and normalize each image channel (using the channel statistics from the train set).
    \item For \svhn, we train for 2200 epochs with batch sizes of 50 (training for 240 epochs with batch sizes of 50 reaches within 1\% of the verified error rate).
          The total number of training steps is 100K.
          The rest of the schedule is identical to \cifar.
          During training, we add random translations, and normalize each image channel (using the channel statistics from the train set).
    \item For \imagenet, we train for 160 epochs with batch sizes of 1024.
          The total number of training steps is 200K.
          We decay the learning rate by 10$\times$ at steps 120K and 180K.
          We use warm-up and ramp-up durations of 10K and 100K steps, respectively.
          We use images downscaled to $64 \times 64$ (resampled using pixel area relation, which gives moir\'e-free images).
          During training, we use random crops of $56 \times 56$ and random flips.
          During testing, we use a central $56 \times 56$ crop.
          We also normalize each image channel (using the channel statistics from the train set).
\end{itemize}

The networks trained using \citet{wong2018scaling} were trained using the schedule and learning rate proposed by the authors.
For \citet{madry2017towards}, we used a learning rate schedule identical to IBP and, for the inner optimization, adversarial examples are generated by $7$ steps of PGD with Adam~\citep{kingma2014adam} and a learning rate of $10^{-1}$.
Note that our reported results for these two methods closely match or beat published results, giving us confidence that we performed a fair comparison.

\begin{figure}[b]
\centering
\includegraphics[width=0.43\textwidth]{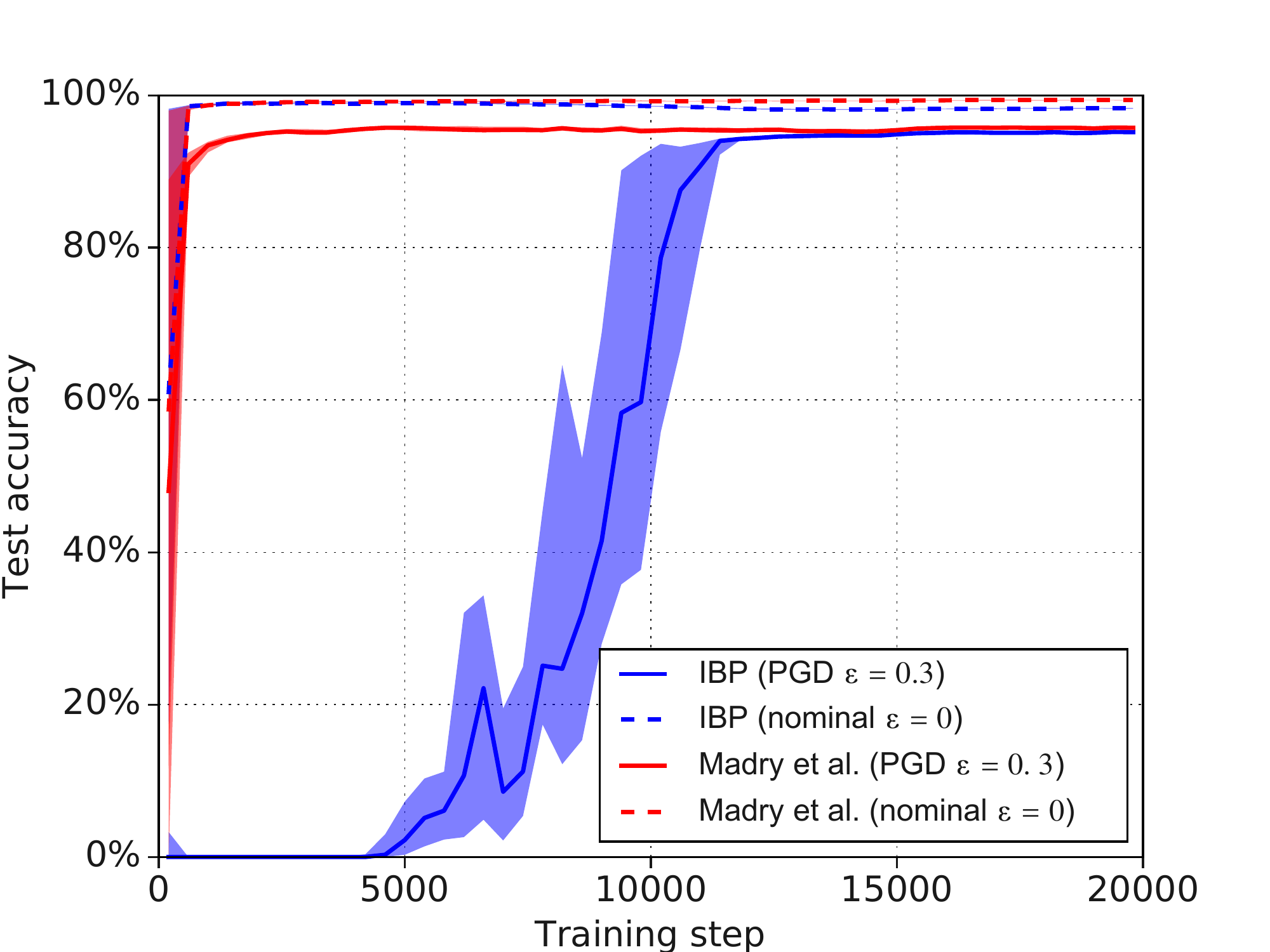}
\caption{Median evolution of the nominal (no attacks) and empirical PGD accuracy (under perturbations of $\epsilon = 0.3$) as training progresses for 10 independently trained large models on \mnist. The shaded areas indicate the 25\textsuperscript{th} and 75\textsuperscript{th} percentiles.}
\label{fig:training_curve}
\end{figure}

Figure~\ref{fig:training_curve} shows how the empirical PGD accuracy (on the test set) increases as training progresses for IBP and \citeauthor{madry2017towards}.
This plot shows the median performance (along with the 25\textsuperscript{th} and 75\textsuperscript{th} percentiles across 10 independent training processes) and confirms that IBP is stable and produces consistent results.
Additionally, for IBP, we clearly see the effect of ramping the value of $\epsilon$ up during training (which happens between steps 2K and 12K).

\clearpage
\section{Convolutional filters}
\label{appendix:conv}

Figure~\ref{fig:conv_filters} shows the first layer convolutional filters resulting from training a small robust model on \mnist against a perturbation radius of $\epsilon = 0.1$.
Overall, the filters tend to be extremely sparse -- at least when compared to the filters obtained by training a nominal non-robust model (this observation is consistent with \citep{wong2018scaling}).
We can qualitatively observe that \citeauthor{wong2018scaling} produces the sparsest set of filters.

Similarly, as shown in Figure~\ref{fig:conv_filters_cifar10}, robust models trained on \cifar exhibit high levels of sparsity in their convolutional filters. \citeauthor{madry2017towards} seems to produce more meaningful filters, but they remain sparse compared to the non-robust model.

This analysis suggests that IBP strongly limits the capacity of the underlying network.
As a consequence, larger models are often preferable.
Larger models, however, can lead to the explosion of intermediate interval bounds -- and this is the main reason why values of $\epsilon$ must be carefully scheduled.
Techniques that combine tighter (but slower) relaxations with IBP could be used in the initial training stages when training deeper and wider models.

\begin{figure}[b]
\centering
\begin{subfigure}{0.48\textwidth}
\includegraphics[width=\linewidth, trim=2.5cm 1.3cm 2cm 1.3cm, clip]{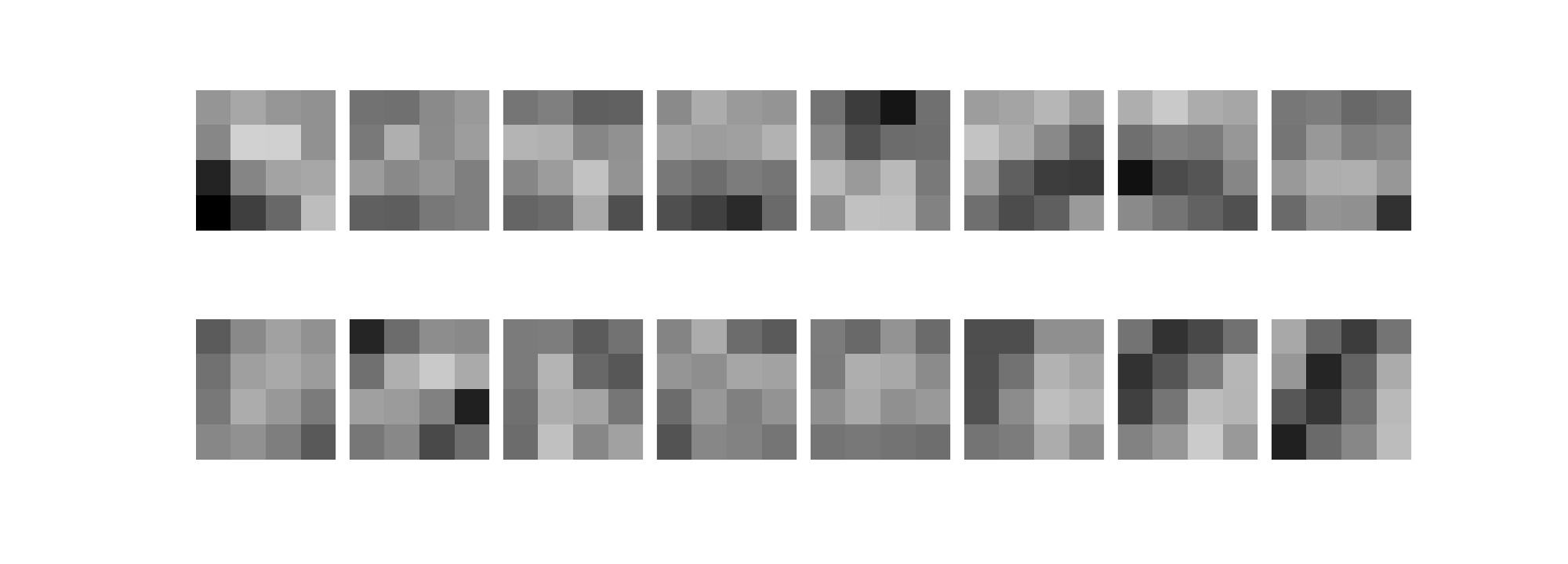}
\caption{Nominal\label{fig:conv_filters_nominal}}
\end{subfigure} \hspace{.3cm}
\begin{subfigure}{0.48\textwidth}
\includegraphics[width=\linewidth, trim=2.5cm 1.3cm 2cm 1.3cm, clip]{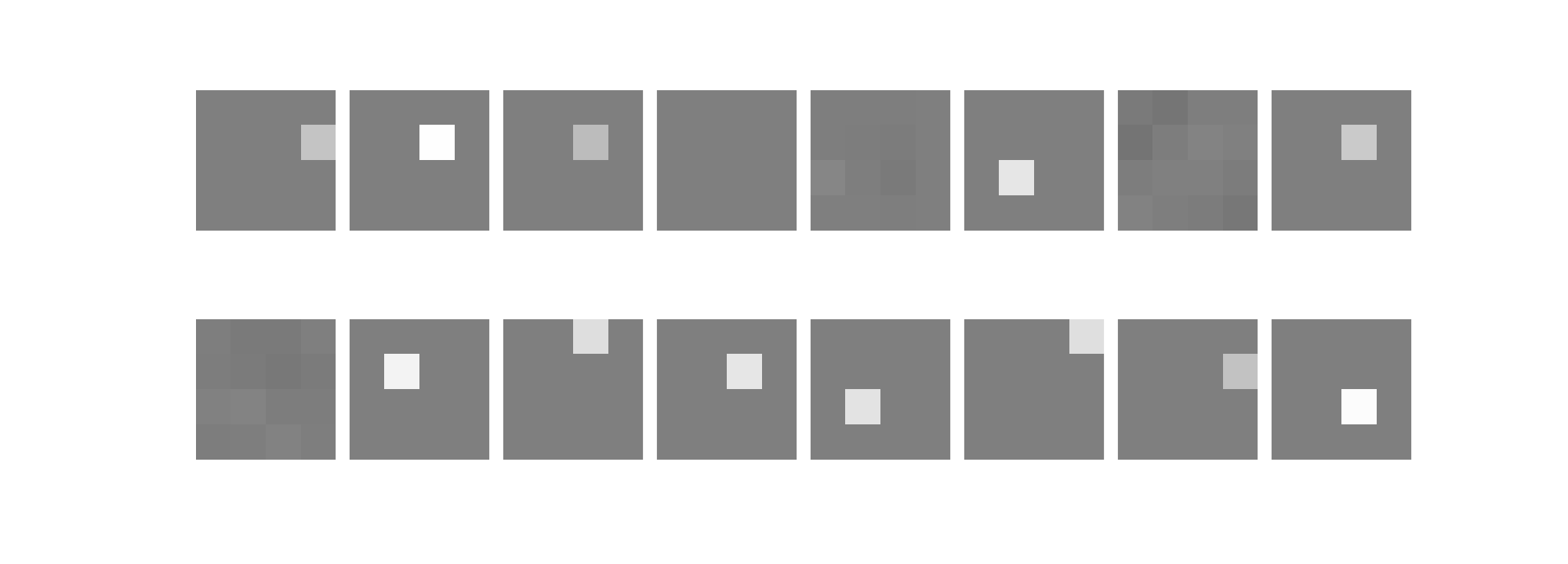}
\caption{IBP\label{fig:conv_filters_naive}}
\end{subfigure} \\
\begin{subfigure}{0.48\textwidth}
\vspace{.5cm}
\includegraphics[width=\linewidth, trim=2.5cm 1.3cm 2cm 1.3cm, clip]{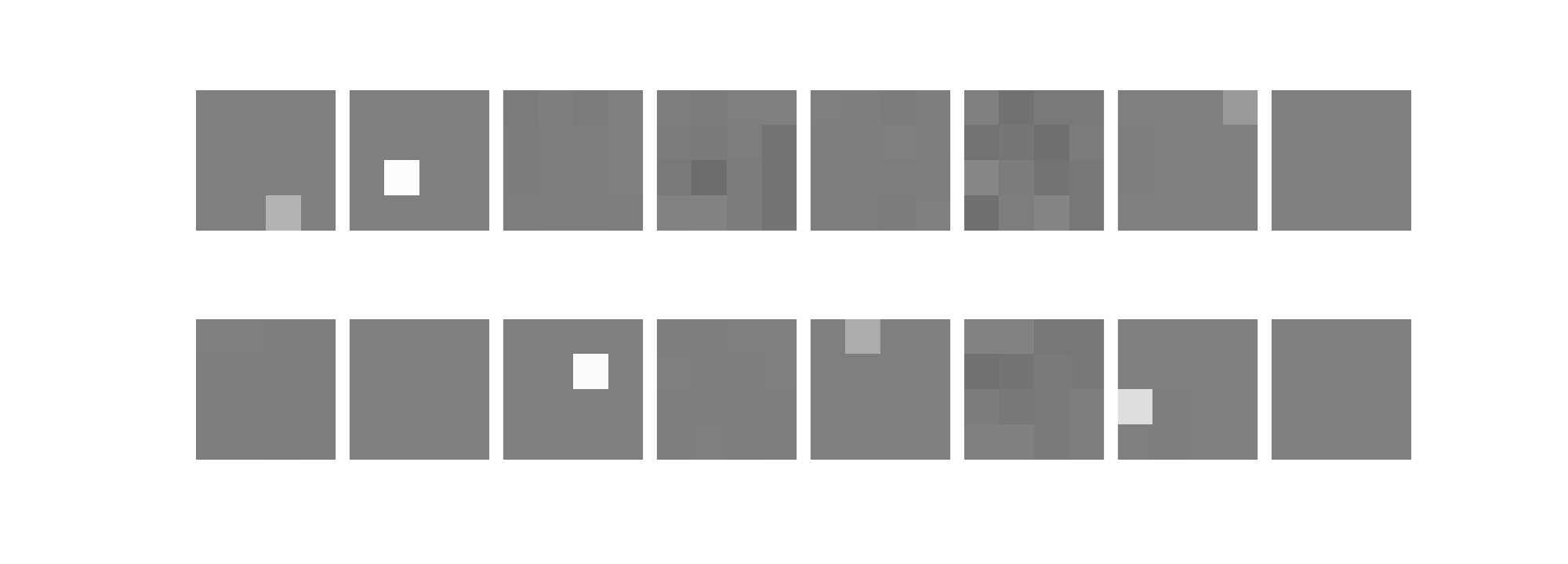}
\caption{\citeauthor{wong2018scaling}\label{fig:conv_filters_wong}}
\end{subfigure} \hspace{.3cm}
\begin{subfigure}{0.48\textwidth}
\vspace{.5cm}
\includegraphics[width=\linewidth, trim=2.5cm 1.3cm 2cm 1.3cm, clip]{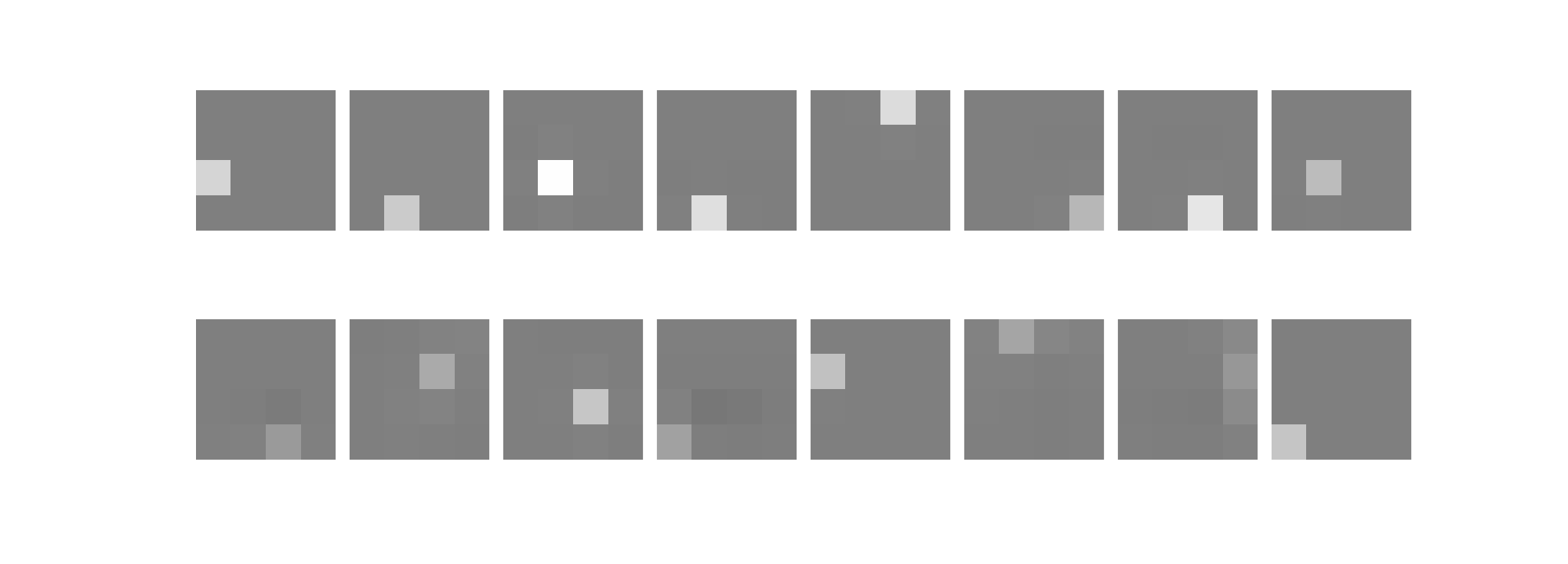}
\caption{\citeauthor{madry2017towards}\label{fig:conv_filters_madry}}
\end{subfigure}
\caption{First layer convolutional filters resulting from training a small robust model on \mnist against a perturbation radius of $\epsilon = 0.1$ for all methods.}
\label{fig:conv_filters}
\end{figure}

\begin{figure}[b]
\centering
\begin{subfigure}{0.48\textwidth}
\includegraphics[width=\linewidth, trim=2.5cm 1.3cm 2cm 1.3cm, clip]{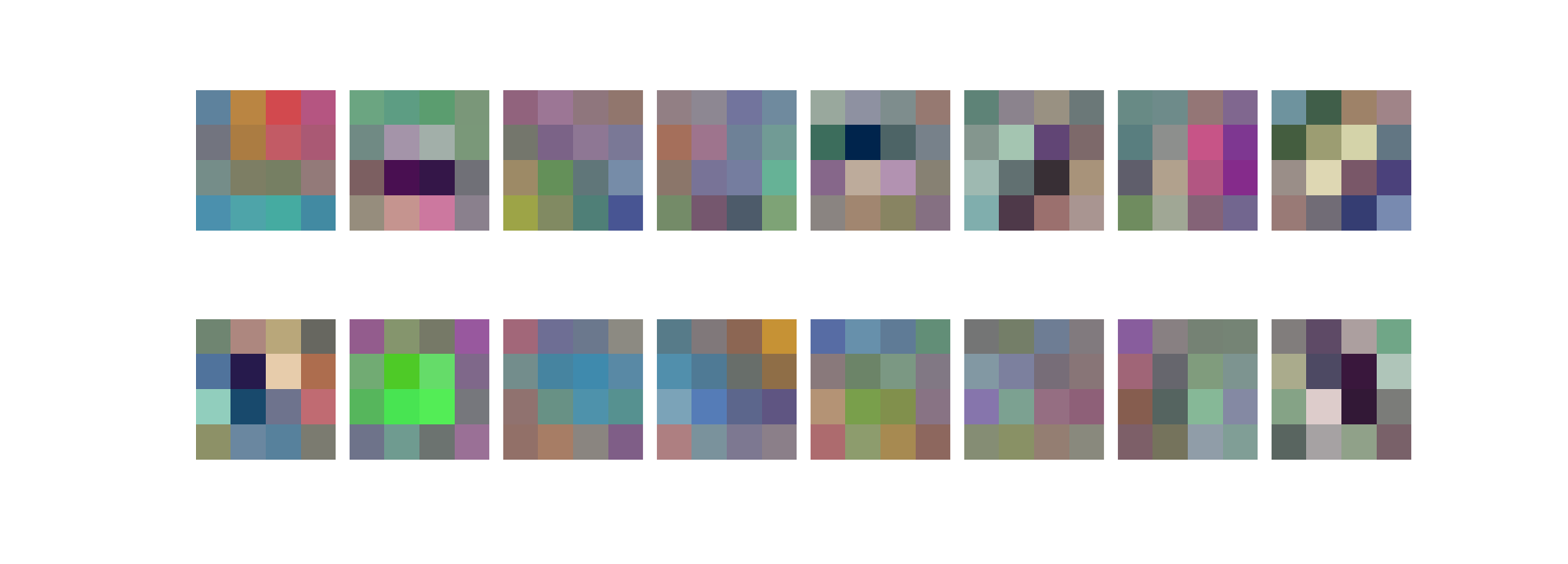}
\caption{Nominal\label{fig:conv_filters_nominal_cifar10}}
\end{subfigure} \hspace{.3cm}
\begin{subfigure}{0.48\textwidth}
\includegraphics[width=\linewidth, trim=2.5cm 1.3cm 2cm 1.3cm, clip]{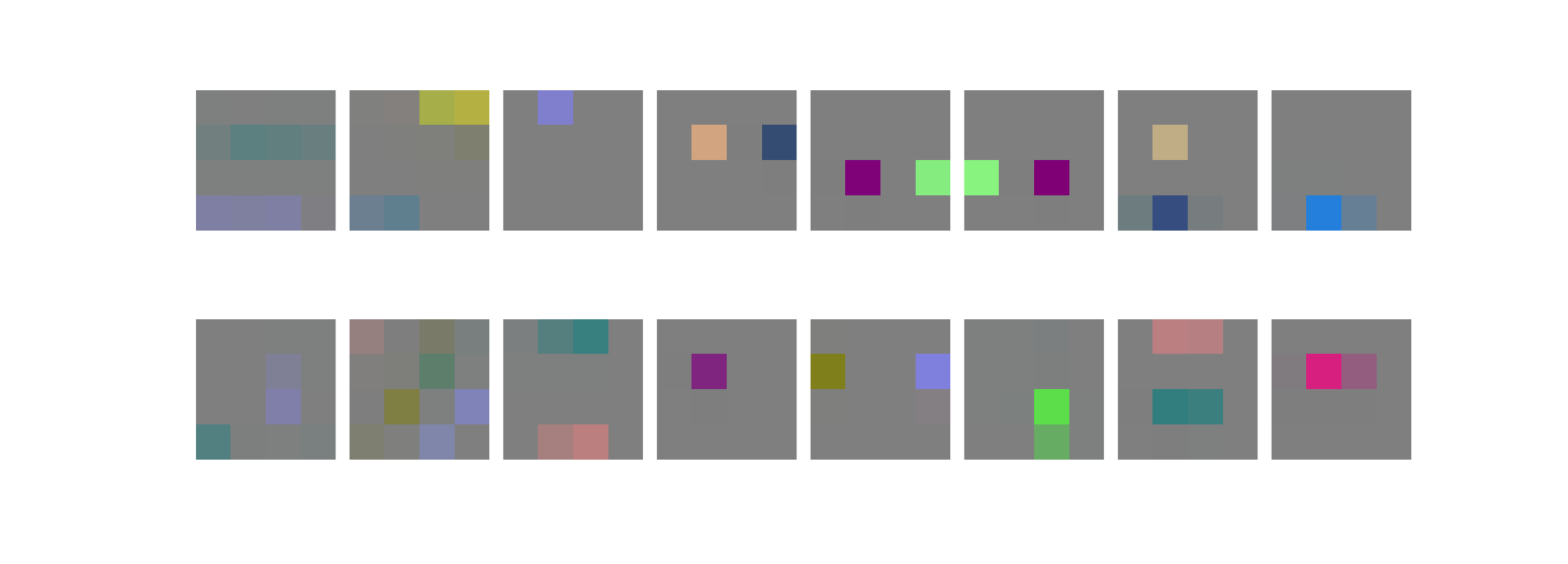}
\caption{IBP\label{fig:conv_filters_naive_cifar10}}
\end{subfigure} \\
\begin{subfigure}{0.48\textwidth}
\vspace{.5cm}
\includegraphics[width=\linewidth, trim=2.5cm 1.3cm 2cm 1.3cm, clip]{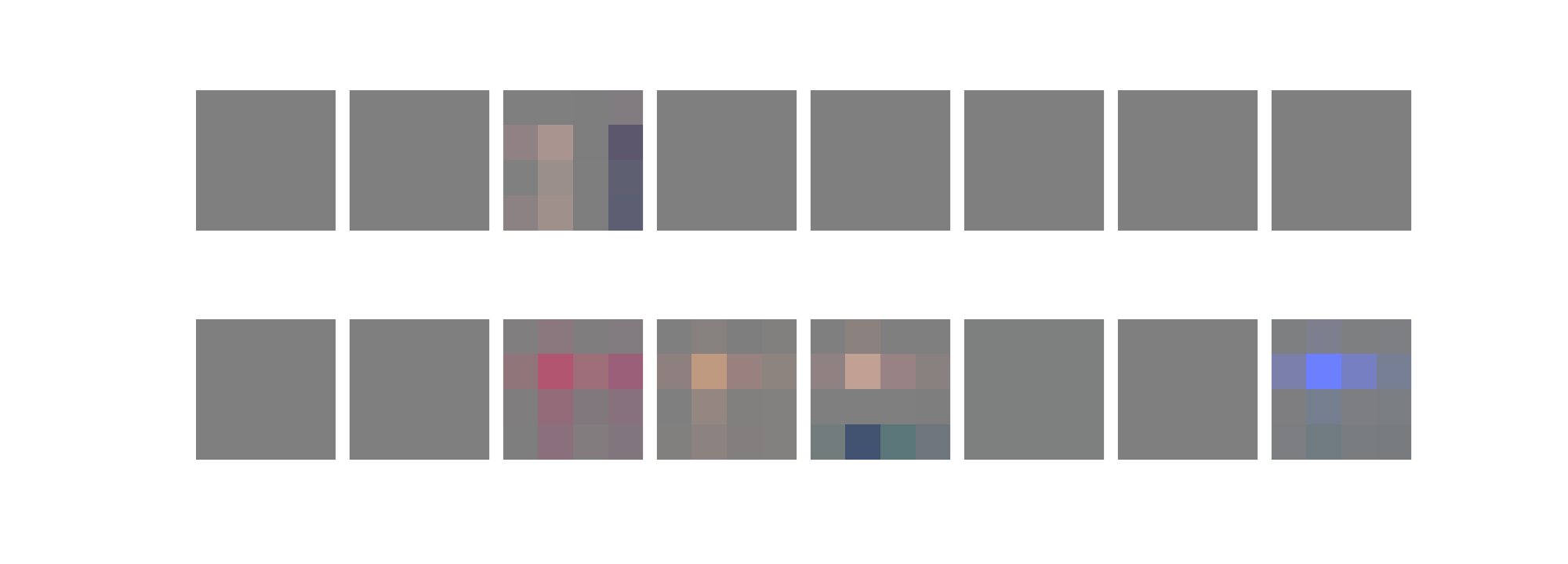}
\caption{\citeauthor{wong2018scaling}\label{fig:conv_filters_wong_cifar10}}
\end{subfigure} \hspace{.3cm}
\begin{subfigure}{0.48\textwidth}
\vspace{.5cm}
\includegraphics[width=\linewidth, trim=2.5cm 1.3cm 2cm 1.3cm, clip]{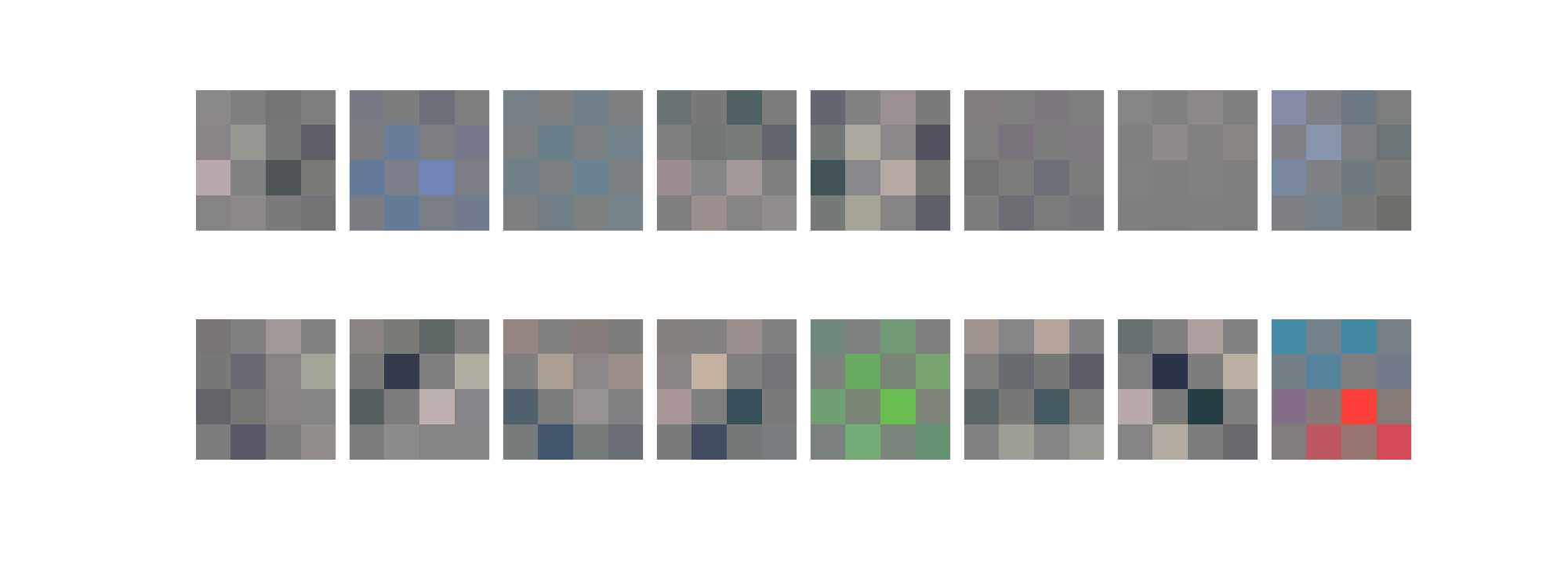}
\caption{\citeauthor{madry2017towards}\label{fig:conv_filters_madry_cifar10}}
\end{subfigure}
\caption{First layer convolutional filters resulting from training a small robust model on \cifar against a perturbation radius of $\epsilon = 2/255$ for all methods.}
\label{fig:conv_filters_cifar10}
\end{figure}

\clearpage
\section{Ablation study}
\label{appendix:ablation}

Table~\ref{table:ablation} reports the performance of each model for different training methods on \mnist with $\epsilon = 0.4$ (averaged over 10 independent training processes).
We chose \mnist for ease of experimentation and a large perturbation radius to stress the limit of IBP.
The table shows the effect of the elision of the last layer and the cross-entropy loss independently.
We do not report results without the schedule on $\epsilon$ as all models without the schedule are unable to train (i.e., reaching only 11.35\% accuracy at best).

We observe that all components contribute to obtaining a good final verified error rate.
This is particularly visible for the small model, where adding elision improves the bound by 3.88\% (12.9\% relative improvement) when using a softplus loss.
Using cross-entropy (instead of softplus) also results in a significant improvement of 4.42\% (14.7\% relative improvement).
Ultimately, the combination of cross-entropy and elision shows an improvement of 5.15\%.
Although present for larger models, these improvements are less visible.
This is another example of how models with larger capacity are able to adapt to get tight bounds.
Finally, it is worth noting that elision tends to provide results that are more consistent (as shown by the 25th and 75th percentiles).

\begin{table}[b]
\begin{center}
\begin{footnotesize}
\begin{tabular}{l|lccc}
    \hline
    \multirow{2}{*}{\bf Model} & \multirow{2}{*}{\bf Training method} & \multicolumn{3}{c}{\bf IBP verified bound} \\
    & & median (0.5) & 0.25 & 0.75 \\
    \hline
    \multirow{4}{*}{small}
    & + $\epsilon$-schedule & 30.09\% & 28.63\% & 32.03\% \\
    & \hspace{.5cm} + elision & 26.21\% & 24.99\% & 28.49\% \\
    & \hspace{.5cm} + cross-entropy & 25.67\% & 24.04\% & 26.71\% \\
    & \hspace{1cm} + elision & \textbf{24.94\%} & \textbf{23.10\%} & \textbf{26.63\%} \\
    \hline
    \multirow{4}{*}{medium}
    & + $\epsilon$-schedule & 20.24\% & 19.77\% & 20.66\% \\
    & \hspace{.5cm} + elision & 19.96\% & 19.92\% & 20.19\% \\
    & \hspace{.5cm} + cross-entropy & 18.10\% & \textbf{17.60\%} & 18.38\% \\
    & \hspace{1cm} + elision & \textbf{18.02\%} & 17.61\% & \textbf{18.11\%} \\
    \hline
    \multirow{4}{*}{large}
    & + $\epsilon$-schedule & 18.32\% & 17.88\% & 18.55\% \\
    & \hspace{.5cm} + elision & 18.03\% & 17.93\% & 18.58\% \\
    & \hspace{.5cm} + cross-entropy & 16.39\% & \textbf{15.73\%} & 17.11\% \\
    & \hspace{1cm} + elision & \textbf{16.33\%} & 16.00\% & \textbf{16.74\%} \\
    \hline
\end{tabular}
\end{footnotesize}
\end{center}
\caption{
{\bf Ablation study.}
This table compares the median verified bound on the error rate (obtained using IBP) for different training methods on \mnist with $\epsilon = 0.4$.
The reported IBP verified error bound is an upper bound of the true verified error rate; it is computed using the elision of the last layer.
The different training methods include combining the linear schedule on \epsilontrain (as explained in Appendix~\ref{appendix:params}) with the elision of the last layer or the cross-entropy loss (the cross-entropy loss is replaced with a softplus loss otherwise, as done by \citealp{mirman2018differentiable} and \citealp{dvijotham2018training}).
We do not report results without the $\epsilon$-schedule as all models without the schedule are unable to train (i.e., reaching only 11.35\% accuracy at best).
}
\label{table:ablation}
\end{table}

\section{Additional results}
\label{appendix:additional_results}

Table~\ref{table:additional_results} provides complementary results to the ones in Table~\ref{table:full_results}.
It includes results for each individual model architecture, as well as results from the literature (for matching model architectures).
The test error corresponds to the test set error rate when there is no adversarial perturbation.
For models that we trained ourselves, the PGD error rate is calculated using 200 iterations of PGD and 10 random restarts.
The verified bound on the error rate is obtained using the MIP/LP cascade described in Section~\ref{sec:results}.

We observe that with the exception of \cifar with $\epsilon = 2/255$, IBP outperforms all other models by a significant margin (even for equivalent model sizes).
For \cifar with $\epsilon = 2/255$, it was important to increase the model size to obtain competitive results.
However, since IBP is cheap to run (i.e., only two additional passes), we can afford to run it on much larger models.

\section{Runtime}

When training the small network on \mnist with a Titan Xp GPU (where standard training takes 1.5 seconds per epoch), IBP only takes 3.5 seconds per epoch compared to 8.5 seconds for \citeauthor{madry2017towards} and 2 minutes for \citeauthor{wong2018scaling} (using random projection of 50 dimensions).
Indeed, as detailed in Section~\ref{sec:nbp} (under the paragraph ``interval bound propagation''), IBP creates only two additional passes through the network compared to \citeauthor{madry2017towards} for which we used seven PGD steps during training.
\citeauthor{xiao2018training}'s method takes the same amount of time as IBP as it needs to perform bound propagation too.

\begin{table*}[tb]
\begin{center}
\begin{footnotesize}
% dataset, epsilon, model, nominal error rate, lower bound, upper bound.
\begin{tabular}{ll|lrrr}
    \hline
    {\bf Dataset} & {\bf Epsilon} & {\bf Method} & {\bf Test error} & {\bf PGD} & {\bf Verified} \\
    \hline
    \multirow{10}{*}{\mnist} & \multirow{10}{*}{$\epsilon = 0.1$}
      & IBP (small) & 1.39\% & 2.91\% & \textbf{2.97\%} \\
    & & IBP (medium) & 1.06\% & 2.11\% & \textbf{2.23\%} \\
    & & IBP (large) & 1.07\% & 1.89\% & 2.32\% \\
    \cline{3-6}
    & & \multicolumn{4}{l}{{\bf Reported in literature}} \\
    & & \hspace{.5cm}\citet{wong2018scaling} (small) & 1.26\% & -- & 4.48\% \\
    & & \hspace{.5cm}\citet{wong2018scaling} (medium) & 1.08\% & -- & 3.67\% \\
    & & \hspace{.5cm}\citet{mirman2018differentiable}* (small) & 2.4\% & 4.4\% & 5.8\% \\
    & & \hspace{.5cm}\citet{mirman2018differentiable} (medium) & 1.0\% & 2.4\% & 3.4\% \\
    & & \hspace{.5cm}\citet{wang2018mixtrain}** (small) & 1.5\% & 3.7\% & 8.4\% \\
    & & \hspace{.5cm}\citet{wang2018mixtrain} (medium) & 0.5\% & 1.8\% & 4.8\% \\
    
    \hline
    \multirow{10}{*}{\mnist} & \multirow{10}{*}{$\epsilon = 0.3$}
      & IBP (small) & 4.59\% & 9.09\% & \textbf{10.25\%} \\
    & & IBP (medium) & 2.70\% & 7.98\% & \textbf{9.74\%} \\
    & & IBP (large) & 1.66\% & 6.12\% & 8.05\% \\
    \cline{3-6}
    & & \multicolumn{4}{l}{{\bf Reported in literature}} \\
    & & \hspace{.5cm}\citet{wong2018scaling} (small) & 14.87\% & -- & 43.10\% \\
    & & \hspace{.5cm}\citet{wong2018scaling} (medium) & 12.61\% & -- & 45.66\% \\
    & & \hspace{.5cm}\citet{mirman2018differentiable} (small) & 3.2\% & 9.0\% & 19.4\% \\
    & & \hspace{.5cm}\citet{mirman2018differentiable} (medium) & 3.4\% & 6.2\% & 18.0\% \\
    & & \hspace{.5cm}\citet{wang2018mixtrain} (small) & 4.6\% & 12.7\% & 48\% \\
    & & \hspace{.5cm}\citet{wang2018mixtrain} (medium) & 3.4\% & 10.6\% & 41.6\% \\
    
    \hline
    \multirow{3}{*}{\cifar} & \multirow{3}{*}{$\epsilon \approx 1/510$***}
      & \multicolumn{4}{l}{{\bf Reported in literature}} \\
    & & \hspace{.5cm}\citet{mirman2018differentiable}*** (small) & 42.8\% & 46.4\% & 47.8\% \\
    & & \hspace{.5cm}\citet{mirman2018differentiable} (medium) & 52.8\% & 59.2\% & 64.2\% \\
    
    \hline
    \multirow{10}{*}{\cifar} & \multirow{10}{*}{$\epsilon \approx 2/255$***\textsuperscript{(}*\textsuperscript{)}}
      & IBP (small) & 39.54\% & 53.95\% & 56.43\% \\
    & & IBP (medium) & 37.28\% & 51.73\% & 54.78\% \\
    & & IBP (large) & 29.84\% & 45.09\% & 49.98\% \\
    \cline{3-6}
    & & \multicolumn{4}{l}{{\bf Reported in literature}} \\
    & & \hspace{.5cm}\citet{wong2018scaling} (small) & 38.91\% & -- & \textbf{52.75\%} \\
    & & \hspace{.5cm}\citet{wong2018scaling} (medium) & 31.28\% & -- & \textbf{46.59\%} \\
    & & \hspace{.5cm}\citet{mirman2018differentiable}*** (small) & 45.8\% & 60.0\% & 64.8\% \\
    & & \hspace{.5cm}\citet{mirman2018differentiable} (medium) & 51.6\% & 61.4\% & 75.8\% \\
    & & \hspace{.5cm}\citet{wang2018mixtrain}**** (small) & 28.9\% & 45.4\% & 62.4\% \\
    & & \hspace{.5cm}\citet{wang2018mixtrain} (medium) & 22.1\% & 36.5\% & 58.4\% \\
    
    \hline
    \multirow{6}{*}{\cifar} & \multirow{6}{*}{$\epsilon = 8/255$}
      & IBP (small) & 61.63\% & 70.42\% & \textbf{72.93\%} \\
    & & IBP (medium) & 58.23\% & 69.72\% & \textbf{72.33\%} \\
    & & IBP (large) & 50.51\% & 65.23\% & 67.96\% \\
    \cline{3-6}
    & & \multicolumn{4}{l}{{\bf Reported in literature}} \\
    & & \hspace{.5cm}\citet{wong2018scaling} (small) & 72.24\% & -- & 79.25\% \\
    & & \hspace{.5cm}\citet{wong2018scaling} (medium) & 80.56\% & -- & 83.43\% \\
    
    \hline
\end{tabular}
\end{footnotesize}
\end{center}
\caption{
{\bf Additional comparison with the state-of-the-art.}
Comparison of the nominal test error (no adversarial perturbation), error rate under PGD attacks, and verified bound on the error rate.
For IBP models, the PGD error rate is calculated using 200 iterations of PGD and 10 random restarts. For reported results, we report the closest setting to ours.
We indicate in parenthesis the model architecture: for \citet{wong2018scaling} small and medium are equivalent to ``Small'' and ``Large''; for \citet{mirman2018differentiable} they are equivalent to the best of ``ConvSmall/ConvMed'' and best of ``ConvBig/ConvSuper''; for \citet{wang2018mixtrain} they are equivalent to ``Small'' and ``Large'' (we always report the best verified result when different verification methods are used). We indicate in bold the best results per model architecture.\\
* \citet{mirman2018differentiable} report results on 500 samples for both \mnist and \cifar (instead of the 10K samples in test set). \\
** The number of samples used in \citet{wang2018mixtrain} is unknown.\\
*** We confirmed with \citeauthor{mirman2018differentiable} that \citep{mirman2018differentiable} uses a perturbation radius of 0.007 post-normalization, it is roughly equivalent to 1/510 pre-normalization. Similarly they use a perturbation radius of 0.03 post-normalization, it is roughly equivalent to 2/255 pre-normalization.\\
**** We confirmed with \citeauthor{wang2018mixtrain} that \citep{wang2018mixtrain} use a perturbation radius of 0.0348 post-normalization, it is roughly equivalent to 2/255 pre-normalization.
}
\label{table:additional_results}
\end{table*}

\clearpage
\section{When Projected Gradient Descent is not enough}
\label{appendix:pgd_vs_mip}

For a given example in \mnist, this section compares the worst-case attack found by PGD with the one found using a complete solver.
The underlying model is a medium sized network trained using IBP with $\epsilon = 0.1$.
The nominal image, visible in Figure~\ref{fig:nominal_image}, has the label ``eight'', and corresponds to the 1365\textsuperscript{th} image of the test set.

The worst-case perturbation of size $\epsilon = 0.1$ found using 200 PGD iterations and 10 random restarts is shown in Figure~\ref{fig:pgd_image}.
In this particular case, the robust network is still able to successfully classify the attack as an ``eight''.
Without any verifiable proof, we could wrongly assume that our network is robust to \linf perturbation on that image.
However, when running a complete solver (using a MIP formulation), we are able to find a counter-example that successfully induces the model to misclassify the ``eight'' as a ``two'' (as shown in Figure~\ref{fig:mip_image}).

\begin{figure}[h]
\centering
\begin{subfigure}{0.2\textwidth}
\centering
\includegraphics[height=2cm, trim=1cm 0.5cm 9.5cm 0.5cm, clip]{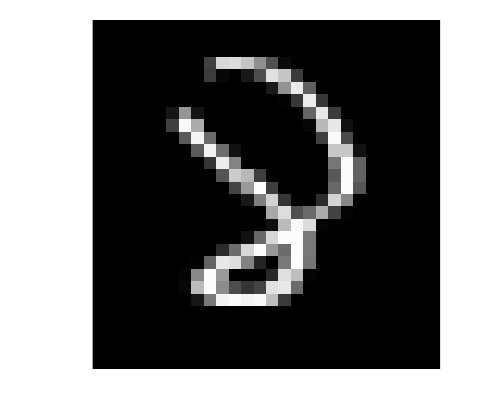}
\caption{Nominal image correctly classified as an ``eight''\label{fig:nominal_image}}
\end{subfigure} \hspace{.5cm}
\begin{subfigure}{0.35\textwidth}
\centering
\includegraphics[height=2cm, trim=6cm 0.5cm 1cm 0.5cm, clip]{images/image_mnist.pdf}
\caption{Worst attack found using PGD still classified as an ``eight''\label{fig:pgd_image}}
\end{subfigure} \hspace{.5cm}
\begin{subfigure}{0.35\textwidth}
\centering
\includegraphics[height=2cm, trim=6cm 0.5cm 1cm 0.5cm, clip]{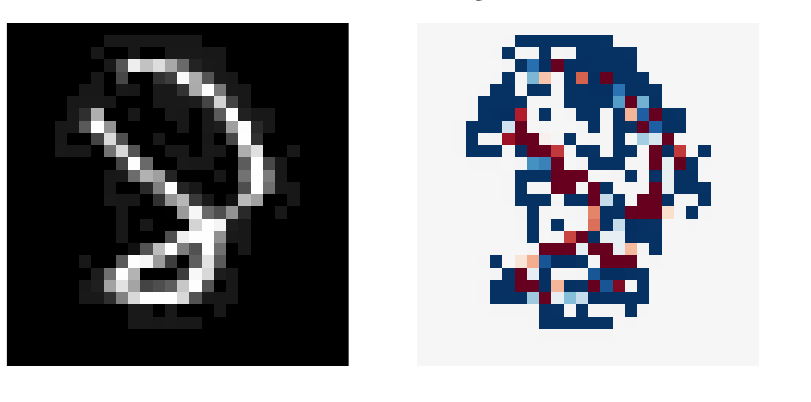}
\caption{Actual worst attack found using a MIP solver incorrectly classified as a ``two''\label{fig:mip_image}}
\end{subfigure}
\caption{Attacks of size $\epsilon = 0.1$ found on the 1365\textsuperscript{th} image of the \mnist test set. For (\subref{fig:pgd_image}) and (\subref{fig:mip_image}), the left pane shows the adversarial image, while the right pane shows the perturbation rescaled for clarity.}
\label{fig:image}
\end{figure}

Figure~\ref{fig:loss_landscape} shows the untargeted adversarial loss (optimized by PGD) around the nominal image.
In these loss landscapes, we vary the input along a linear space defined by the worse perturbations found by PGD and the MIP solver.
The $u$ and $v$ axes represent the magnitude of the perturbation added in each of these directions respectively and the $z$ axis represents the loss.
Typical cases where PGD is not optimal are often a combination of two factors that are qualitatively visible in this figure:
\begin{itemize}
    \item We can observe that the MIP attack only exists in a corner of the projected \linf-bounded ball around the nominal image.
    Indeed, since PGD is a gradient-based method, it relies on taking gradient steps of a given magnitude (that depends on the learning rate) at each iteration.
    That is, unless we allow the learning rate to decay to a sufficiently small value, the reprojection on the norm-bounded ball at each iteration will force the PGD solution to bounce between the edges of that ball without hitting its corner.
    \item The second, more subtle, effect concerns the gradient direction.
    Figure~\ref{fig:loss_landscape_2d}, which shows a top-view of the loss landscape, indicates that a large portion of \linf ball around the nominal image pushes the PGD solution towards the right (rather than the bottom).
    In other words, gradients cannot always be trusted to point towards the true worst-case attack.
\end{itemize}

\begin{figure}[t]
\centering
\begin{subfigure}{0.4\textwidth}
\includegraphics[width=\linewidth, trim=2.2cm 0cm 0cm 1.3cm, clip]{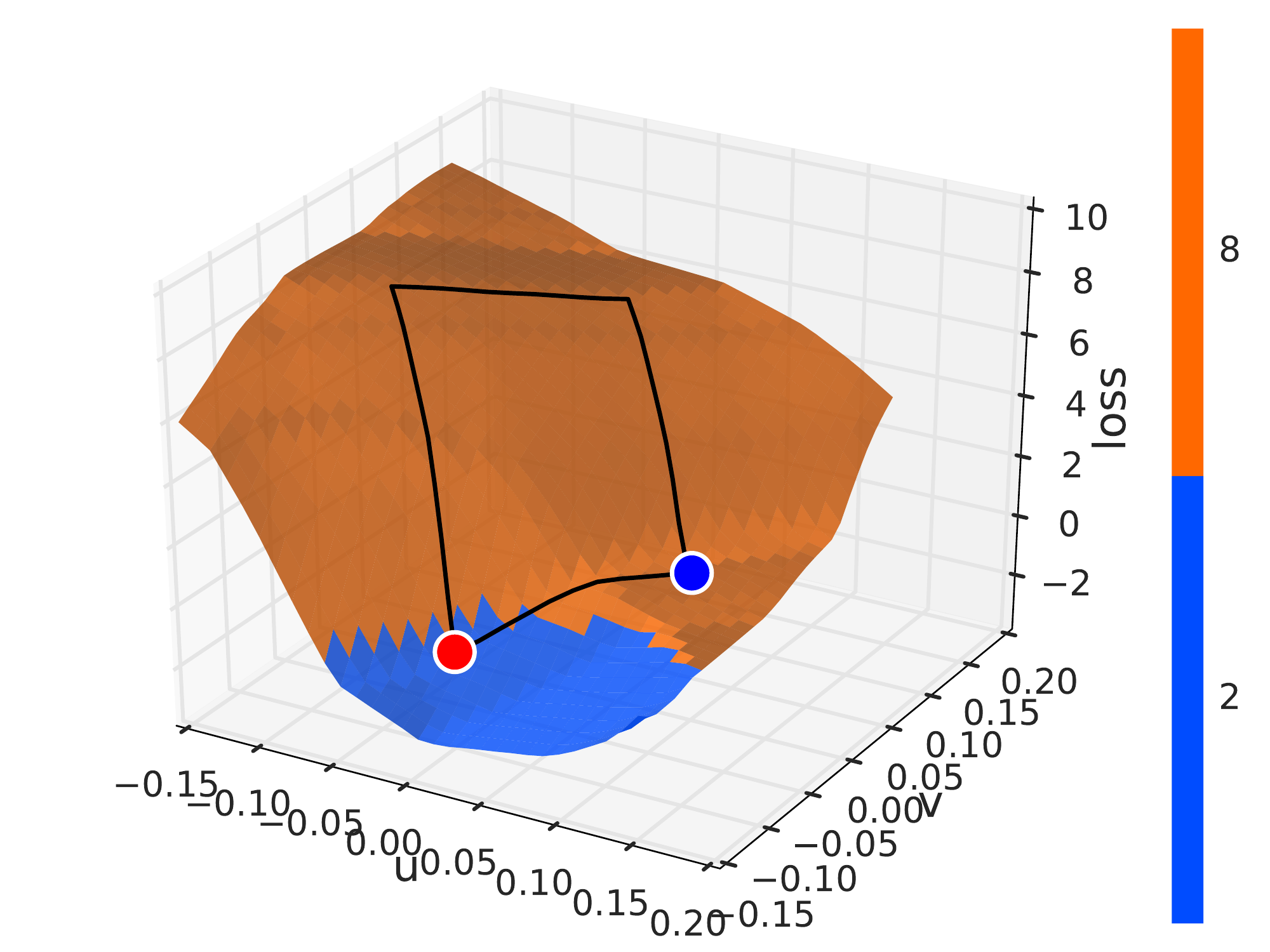}
\caption{\label{fig:loss_landscape_3d}}
\end{subfigure} \hspace{.5cm}
\begin{subfigure}{0.42\textwidth}
\includegraphics[width=\linewidth]{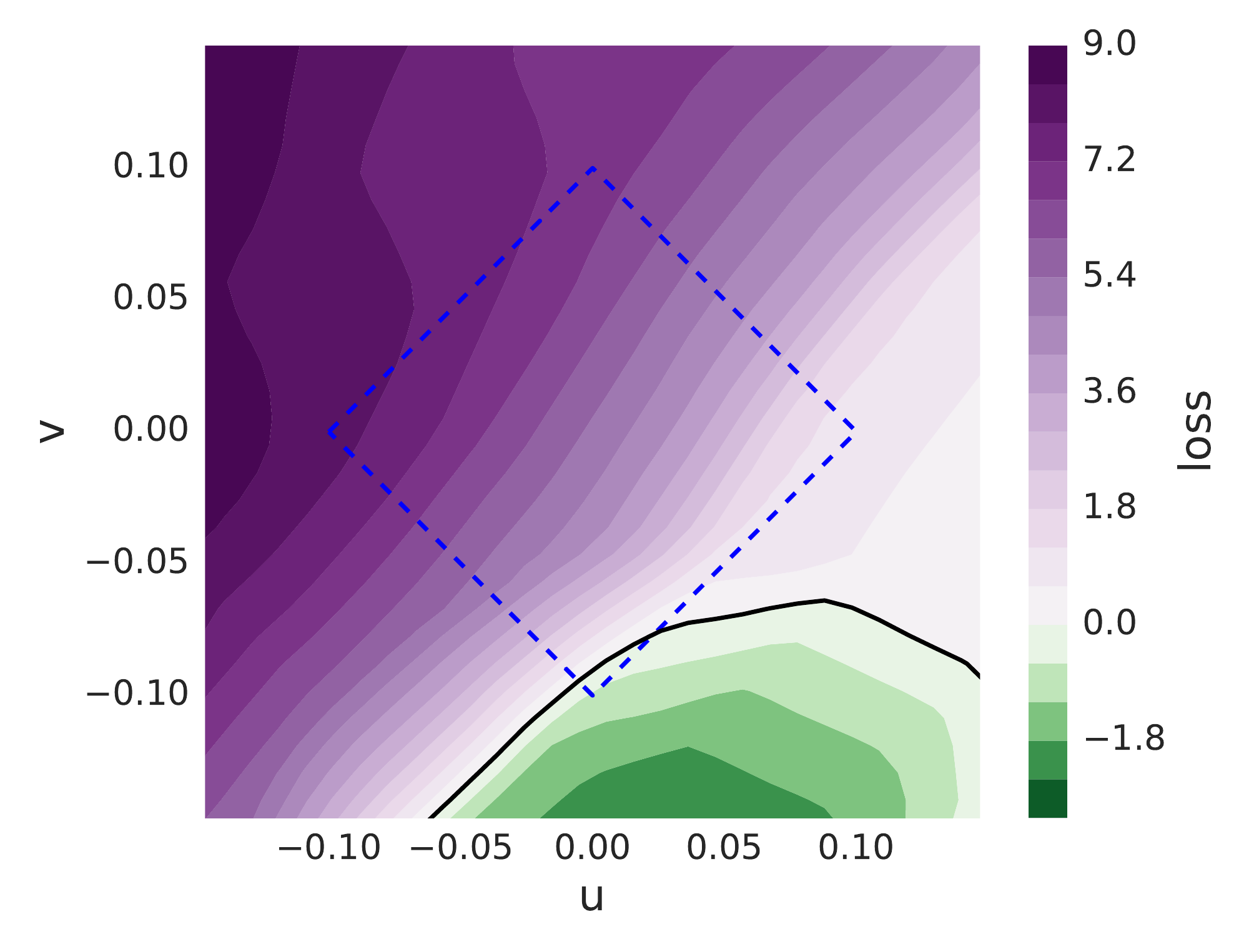}
\caption{\label{fig:loss_landscape_2d}}
\end{subfigure}
\caption{Loss landscapes around the nominal image of an ``eight''.
It is generated by varying the input to the model, starting from the original input image toward either the worst attack found using PGD ($u$ direction) or the one found using a complete solver ($v$ direction). In (\subref{fig:loss_landscape_3d}), the z axis represents the loss and the orange and blue colors on the surface represent the classification predicted by the model. We observe that while the PGD attack (blue dot) is correctly classified as an ``eight'', the MIP attack (red dot) is misclassified as a ``two''. Panel~(\subref{fig:loss_landscape_2d}) shows a top-view of the same landscape with the decision boundary in black. For both panels, the diamond-shape represents the projected \linf ball of size $\epsilon = 0.1$ around the nominal image.}
\label{fig:loss_landscape}
\end{figure}

\fi

\end{document}